%% file: references.tex
\documentclass{article}

\PassOptionsToPackage{numbers, compress}{natbib}
\usepackage[preprint]{neurips_2024}

\usepackage[utf8]{inputenc} % allow utf-8 input
\usepackage[T1]{fontenc}    % use 8-bit T1 fonts
\usepackage{hyperref}       % hyperlinks
\usepackage{url}            % simple URL typesetting
\usepackage{booktabs}       % professional-quality tables
\usepackage{amsfonts}       % blackboard math symbols
\usepackage{nicefrac}       % compact symbols for 1/2, etc.
\usepackage{microtype}      % microtypography
\usepackage{xcolor}         % colors
\usepackage{graphicx}
\usepackage{subcaption}
\usepackage{algorithm}
\usepackage{algpseudocode}

\algrenewcommand{\algorithmiccomment}[1]{\hfill\textcolor{blue}{$\triangleright$ #1}}
\title{Highly Optimized Kernels and Fine-Grained Codebooks for LLM Inference on Arm CPUs}
\author{%
Dibakar Gope $^*$ \qquad David Mansell $^*$ \qquad Danny Loh \qquad Ian Bratt\\
\\
Arm Inc.
\vspace{-2mm}
}

\begin{document}

\maketitle
\def\thefootnote{*}\footnotetext{Algorithm, compression, and quantization contribution: Dibakar Gope \texttt{\{dibakar.gope\}@arm.com}, and CPU kernel optimization contribution: David Mansell \texttt{\{david.mansell\}@arm.com}}\def\thefootnote{\arabic{footnote}}
%Dibakar Gope leads the quantization algorithm, David Mansell leads the kernels optimization.}\def\thefootnote{\arabic{footnote}}
%\def\thefootnote{*}\footnotetext{Correspondence to: Dibakar Gope (\texttt{\{dibakar.gope\}@arm.com}), David Mansell (\texttt{\{david.mansell\}@arm.com}). Dibakar Gope leads the compression and quantization algorithm, David Mansell leads the CPU kernel optimization.}\def\thefootnote{\arabic{footnote}}
\input{sec/abstract}    
\input{sec/intro}
\input{sec/related_work}
\input{sec/motivation}
\input{sec/kernel_optimizations}
\input{sec/nonuniform_quant}
\input{sec/evaluation}

\input{sec/conclusion}
\begin{ack}
% Use unnumbered first level headings for the acknowledgments. All acknowledgments
% More information about this disclosure can be found at: \url{https://neurips.cc/Conferences/2024/PaperInformation/FundingDisclosure}.

% Do {\bf not} include this section in the anonymized submission, only in the final paper. You can use the \texttt{ack} environment provided in the style file to automatically hide this section in the anonymized submission.
We would like to thank Eric Biscondi, Milos Puzovic, Gian Marco Iodice, Ronan Naughton, Ashok Bhat, Ravi Malhotra, Pavan Gorti, Masoud Koleini, Adnan AlSinan, Mariusz Czyz, and Sandeep Singh from
Arm for their insightful discussions, feedback, and support.
\end{ack}

{\small
\bibliographystyle{plainnat}
\bibliography{references}
}

\end{document}

%% file: sec/abstract.tex
\begin{abstract}
Large language models (LLMs) have transformed the way we think about language understanding and generation, enthralling both researchers and developers. However, deploying these models for inference has been a significant challenge due to their unprecedented size and resource requirements. Facilitating the efficient execution of LLMs on commodity Arm CPUs will expand their reach to billions of compact devices such as smartphones and other small devices. 
While quantizing model weights to sub-byte precision (for example, $4$ bits per weight or less) has emerged as a promising solution to ease memory pressure, the group quantization formats commonly used for LLM quantization have significant compute overheads and a resource-intensive dequantization process. As a result, a higher proportion of compute instructions do not perform multiplies, i.e., real work, rendering them unsuitable for meeting the required latency requirements for LLM variants deployed on commodity CPUs. In addition, CPU-based LLM inference has received far less attention in previous efforts. In this work, we propose a set of highly optimized kernels to accelerate LLM inference, demonstrate the best possible performance, and unleash the full potential of CPUs, particularly Arm CPUs. These kernels amortize the cost of loading the operands and the cost of weight unpacking across multiple output rows. This, along with the introduction of an optimized interleaved group data layout format for weights and decompression path optimizations to reduce unnecessary operations and dequantization overhead while maximizing the use of vector and matrix multiply operations, significantly improves the efficiency of MAC operations. Furthermore, we present a group-wise non-uniform codebook-based quantization method for ultra-low-precision quantization of LLMs to better match non-uniform patterns in their weight distributions, allowing large-scale LLMs to fit on smaller devices and demonstrating better throughput during token generation while ensuring better quality than the state-of-the-art. Experiments show that applying these improvements to LLMs with $4$-bit and $2$-bit quantization results in at least $3$-${3.2\times}$ improvement in prompt processing and ${2\times}$ improvement in autoregressive decoding on a single Arm CPU core, compared to LLaMA.cpp-based solution. The optimized kernels are available at \url{https://github.com/ggerganov/llama.cpp}. %\href{https://github.com/ggerganov/llama.cpp}{https://github.com/ggerganov/llama.cpp}.
\end{abstract}

%% file: sec/intro.tex
\section{Introduction}
\label{sec:intro}

Generative Large Language Models (LLMs) have demonstrated remarkable results for a wide range of tasks. A language model can predict the next word given a context or a question. LLMs are trained with massive amounts of data to learn language patterns. They can perform tasks ranging from summarizing and translating texts to responding in chatbot conversations. As a result, facilitating their efficient execution on Arm CPUs will expand their reach to billions of Arm devices. LLMs are often memory-bandwidth and memory capacity-bound, with memory accesses dominated by weights, allowing CPUs the opportunity to achieve competitive performance and outperform other processors and accelerators in terms of overall inference/cost. Furthermore, Arm CPUs are pervasive, providing portability and flexibility, so a new LLM compression scheme can work seamlessly on Arm CPUs without much effort. Given all the advantages, this work attempts to unlock the full potential of LLMs on Arm CPUs deployed in dataceneters, smartphones, and edge devices.

Deploying these LLMs for inference has been a significant challenge due to their unprecedented size and resource requirements. One of the primary performance bottlenecks in LLM inference for generative tasks is memory bandwidth. Quantization has been an effective approach to converting high-precision ($16$ or $32$-bit) model weights to lower-precision values without significantly affecting accuracy. It lowers the model's memory and computational requirements, making it better suited for deployment on devices with limited resources. While $8$-bit quantization reduces LLM storage requirements by half, the large scale size of LLMs necessitates quantizing them to even lower precisions (for example, $4$, or even lower bit-widths). When quantized to $2$ bits, the Llama3 $70$B-like large-scale foundation model requires less than $20$ GB of memory. As a result, there has been a significant research interest in achieving even greater compression through ultra-low-precision (e.g., $2$ bits per weight) quantization, non-uniform quantization, and complex compression schemes.
%This not only allows large models to fit on smaller devices, but it also reduces memory bandwidth pressure during the autoregressive decoding stage, resulting in improved throughput.
While quantization has emerged as a promising solution to the memory bandwidth problem, expensive decoding schemes and large model footprint access of existing quantization methods continue to have a significant impact on runtime performance. 
With this advancement in quantization formats and compression algorithms targeting LLMs, a major obstacle still to be overcome is providing effective system support for these compressed numerical formats for extreme compression regimes ($4$ or fewer bits per weight) so that LLMs can be executed quickly and accurately on end devices such as GPUs or CPUs. 
This motivates the development of faster inference kernels as well as runtime-friendly, fast, and accurate quantization methods.

%With this advancement in quantization formats and compression algorithms targeting LLMs, a major obstacle still to be overcome is providing effective system support for these compressed numerical formats for extreme compression regimes (4 or fewer bits per weight) so that LLMs can be executed quickly and accurately on end devices such as GPUs or CPUs. 
While much of the prior research has focused on GPU-based inference as the primary target scenario~\cite{kwon2023vLLM,lin2024qserve,sheng2023flexgen,kurtic2024givebf16,guo2024flute2024}, CPU-based inference has received significantly less attention. Although open-source CPU-based inference frameworks, such as LLaMA.cpp~\cite{llamacpp}, can offer decent time-to-first-token and generative performance on off-the-shelf CPUs, in this work, we investigate the possibility of achieving significantly better runtime performance by designing optimized matrix-vector multiplications (GEMV) and matrix-matrix multiplications (GEMM) kernels targeting LLM inference of varying low-bitwidths on Arm CPUs.

%Furthermore, existing quantization methods do not perform well and result in notable degradation quality at extreme compression ratios, such as 2-bit quantization. 
Furthermore, existing quantization methods do not perform well at extreme compression ratios, such as $2$-bit quantization, and result in either poor runtime performance despite advanced kernel optimizations or notable degradation in quality.
We then propose a novel non-uniform codebook-based post-training quantization method that enables ultra-low-precision quantization of LLMs while outperforming the state-of-the-art in terms of text generation quality and runtime performance. This is accomplished through the innovation of applying non-uniform codebook-based quantization over group-wise structured LLM weight matrices, fine-grained codebook assignment to weight groups, in conjunction with identifying and using as few codebooks for all LLM layers as possible, so that the codebooks for all of them can be stored in an Arm CPU's register file (for example, a single $128$-bit vector register). This, combined with Arm CPU-optimized codebook-based group-wise quantized matrix multiply kernels, results in significantly improved runtime performance for foundation models in the domain of LLM. While we consider ARM CPU-based generative inference as the motivating setup for our work, our techniques are general and should be extensible to other settings as well.

The key contributions of this work are as follows:
\vspace{-2mm}
\begin{itemize}
%\item \textbf{We consider CPU-based generative inference as the motivating setup for our work, although our techniques are general, and should be extensible to other settings as well.}
%\item In this work, we use the LLaMA2 7B parameter 4b quantized model as a benchmark. We used a state-of-the-art c/c++ runtime such as llama.cpp for our evaluation. However, we realize that the baseline kernels of the C++ runtime cannot exploit the true potential of Arm CPUs. As a result, we developed a set of highly optimized GEMV and GEMM kernels for 4b quantized LLMs and demonstrated significant improvement in the two runtime metrics, such as time-to-first-token and tokens/s that we care about in the context of LLMs.
\item We develop a set of highly optimized GEMV and GEMM kernels for various low bit-width group-wise quantized LLMs. With the help of SIMD-aware weight packing and fast decompression path optimizations, these kernels can fully take advantage of available vector and matrix multiply instructions to maximize MAC unit utilization, minimize overhead and memory accesses, and achieve the best possible performance (to date) on Arm CPUs. %They enable faster throughput in both the compute-bound time-to-first-token (prefill) and the memory-bound token generation (autoregressive decoding) stages of LLM inference when compared to the state-of-the-art.
Our optimized $4$-bit group-wise quantization kernels enable $3$-$3.2\times$ faster throughput in the compute-bound time-to-first-token (prefill) and $2\times$ higher throughput in the memory-bound token generation (autoregressive decoding) stages of LLM inference on Arm CPUs when compared to the state-of-the-art.
%Our optimized kernels also show the effectiveness of various non-uniform quantization methods, such as different scalar and vector quantization types and complex compression schemes. 
We also present highly optimized kernels for various non-uniform quantization methods, such as scalar and vector quantization types, as well as narrower $2$-bit quantization, and demonstrate the effectiveness of off-the-shelf Arm CPUs in offering high throughput performance for them.
%Furthermore, we provide a set of guidelines for developing runtime-friendly quantization methods.
%Our framework significantly speeds up linear layers through on-the-fly dequantization. We also take advantage of efficient 4-bit weight packing and kernel fusion to minimize the inference overhead
\item %We propose a group-wise codebook-based quantization method for ultra-low-precision quantization of LLMs to better match non-uniform patterns in their weight distributions, allowing large-scale LLMs to fit on smaller devices and demonstrating better throughput during token generation while ensuring better quality than the state-of-the-art. 
%We propose a group-wise codebook-based quantization method for ultra-low-precision quantization of LLMs to better match non-uniform patterns in their weight distributions. This allows large-scale LLMs to fit on smaller devices and achieve better quality than the state-of-the-art (about a 0.9-point improvement in perplexity at similar bits per weight for the LLaMA-3 8B model) while demonstrating similar throughput to that of low-decoding overhead uniform quantization techniques during token generation.
We propose a group-wise codebook-based quantization method for ultra-low-precision quantization of LLMs to better match non-uniform patterns in their weight distributions. Our fine-grained non-uniform quantization technique not only achieves better LLM quality than the current state-of-the-art (about a $0.9$-point improvement in perplexity at similar bits per weight for the LLaMA-3 $8$B model) but also demonstrates a high throughput comparable to low-decompression overhead uniform quantization techniques during token generation.
We present a pareto-optimal solution in terms of model quality and runtime performance for low bit-width quantization, outperforming state-of-the-art $2$-$3$-bit quantization techniques in terms of LLM quality while requiring similar bits per weight and no additional finetuning.
% and outperform the state-of-the-art 2-bit quantization techniques in terms of LLM quality while requiring %fewer
% similar bits per weight and no additional training or finetuning.
\vspace{-2mm}
%\item Develops a constraint-satisfaction-guided clustering algorithm to limit the number of codebooks to a few (for example, 1, 2, 4, etc.), each with a few entries (for example, 4, 8, or 16 depending on the quantization bit width).
%\item Does not require to load the codebook entries to the register file for different weight rows or LLM layers; the register file (for example, a single 128-bit vector register) can store the entire codebook once it has been loaded from memory once.
%\item The same codebook works across all layers of an LLM and across all LLMs; there is no need to determine the codebook entries (centroid values in a codebook) given a new unseen LLM.
%\item Requires no additional training or finetuning.
%\item Develops Arm CPU architecture-optimized matrix-multiply kernels that can seamlessly use existing table vector lookup instructions vqtbl to take advantage of the new codebook-based quantization method and cope with various low bit width quantizations when reading elements from codebooks.
%\item Enables faster throughput in both the compute-bound time-to-first-token (prefill) and the memory-bound token generation (autoregressive decoding) stages when compared to the state-of-the-art
\end{itemize}

%% file: sec/related_work.tex
\section{Related work}
\label{sec:related_work}
%\textbf{Model quantization methods. }
\textbf{Quantization of LLMs. }
Due to the massive size of LLMs, post-training quantization (PTQ) methods have emerged as an essential technique for accelerating LLMs during inference and running them efficiently. There has been a surge of interest and an increasing body of work in developing accurate PTQ methods targeting LLMs in recent times, as doing so can directly lower the cost of running them. By reducing the precision of pre-trained LLMs, PTQ methods save memory and speed up LLM inference while preserving most of the model quality at scale when compared to the performance and compute requirements of other compression techniques such as pruning and quantization-aware training (QAT). Early PTQ works on LLMs such as ZeroQuant~\citep{yao2022zeroquant}, LLM.int8()~\citep{dettmers2022gptint}, and nuQmm~\citep{park2024lutgemm} demonstrate the potential to use fine-grained quantization (i.e., group-wise quantization) for model weights to achieve better accuracy while at the cost of slightly less compression in comparison to standard coarser-grained quantization methods. Subsequent quantization works, such as GPTQ~\citep{frantar2023optq}, compress LLM weights more aggressively to $3$ or $4$ bits, unlocking the possibility of running massive LLMs on consumer hardware. GPTQ employs layer-wise quantization in conjunction with Optimal Brain Quantization (OBQ)~\citep{frantar2022optimal}, in which the easiest weights are quantized first, and all remaining non-quantized weights are adjusted to compensate for the precision loss. This, combined with fine-grained group-wise quantization, results in high compression rates while maintaining high quality. QuIP~\citep{chee2023quip} and QuIP\#~\citep{tseng2024quip} apply incoherence processing to further quantize LLMs to $2$ bits per weight, recognizing that quantization benefits from incoherent weights and corresponding proxy Hessian matrices. SqueezeLLM~\citep{kim2024sqllm}, AWQ~\citep{lin2024awq}, and SpQR~\citep{dettmers2023spqr} lines of work observe that a small subset of LLM model weights produce noticeably large quantization errors, therefore storing them with higher precision to counteract the accuracy degradation caused by their weight quantization and demonstrating improved PTQ accuracy over previous works. While aggressive weight quantization is critical for LLMs to reduce inference costs, activation quantization is less of an issue due to their smaller memory footprint, so activations are typically quantized to $8$ bits. The presence of activation outliers can sometimes pose a challenge to weight-activation co-quantization, and subsequent works such as SmoothQuant~\citep{xiao2023smoothquant}, Outlier Suppression~\citep{wei2022outlier, wei2023outliersuppressionplus}, and OmniQuant~\citep{shao2024omniquant} addressed this by introducing a per-channel scaling transformation that shifts the quantization difficulty from activations to weights, allowing activations to be quantized to $8$ bits.

In general, most previous LLM PTQ studies have used group-wise quantization, along with some advanced techniques to handle outliers~\cite{zhao2024atom,yuan2023rptq,ashkboos2024quarot,liu2024spinquant,lee2024owq,liu2024qllm}. The open-source inference framework LLaMA.cpp~\cite{llamacpp} also employs group-wise quantization in conjunction with higher precisions for critical layers, such as $4$-bits for the vast majority of layers and higher precisions for a few.

\textbf{System support for low-bit quantized LLMs. }
The majority of the aforementioned quantization techniques focused on GPU inference as their primary target scenario. For group-wise quantized LLMs, GPTQ-for-LLaMA offers $4$-bit (INT4) Triton kernels, while GPTQ provides $3$-bit (INT3) CUDA kernels. vLLM~\citep{kwon2023vLLM} implements optimized CUDA kernels for INT4, INT8, and FP8 data types for both Nvidia and AMD GPUs, and it also enhances memory efficiency by using PagedAttention to manage attention key and value memory effectively. Nvidia's TensorRT-LLM inference library integrates optimized GPU kernels from FasterTransformer and employs tensor parallelism to enable scalable inference across multiple GPUs. FlashAttention~\citep{dao2022flashattention,dao2024flashattention} combines all of the attention operations into a single kernel and tiles the weight matrices into smaller blocks to better fit the small SRAM, reducing the number of memory accesses between GPU high-bandwidth memory (HBM) and GPU on-chip SRAM. LUT-GEMM~\citep{park2024lutgemm} uses lookup tables to perform bitwise computations on GPU CUDA cores.

On the other hand, CPU-based LLM inference and running modes locally on end devices equipped with CPUs have received far less attention~\cite{zhang2024nomadattention}. The open-source inference framework LLaMA.cpp can offer reasonable generative performance on end devices. Our work develops the most optimized kernels for group-quantized LLMs to date, demonstrating significantly improved performance over LLaMA.cpp for a variety of low bit-widths on Arm CPUs.

\textbf{Non-uniform quantization. }
In addition to uniform quantization techniques as mentioned above, post-training non-uniform quantization techniques have recently received a lot of attention in order to better match the non-uniform patterns commonly found in LLM weight distributions.
SqueezeLLM~\citep{kim2024sqllm} applies k-means clustering to LLM weights to closely approximate the non-uniform distribution and encodes the clusters using codebooks. Recent work on GPTVQ~\citep{vanbaalen2024gptvq}, QuIP\#~\citep{tseng2024quip}, and AQLM~\citep{egiazarian2024aqlm} extends the potential of non-uniform quantization to vector quantization, which quantizes a vector of weights together using codebooks and thus captures the shape of the source distribution more accurately than SqueezeLLM-like scalar quantization approaches. However, the high overhead of accessing codebooks, combined with the complex decompression path of the above studies, results in poor runtime performance. In contrast, our group-wise codebook-based quantization ensures not only faster throughput but also better quality than these state-of-the-art approaches under extreme compression scenarios (e.g., $2$-bit quantization).
There are works employing a differentiable k-means approach~\citep{cho2022dkm} as an alternative to performing non-uniform codebook-based quantization on pre-trained models. This approach allows codebooks to be fine-tuned using SGD with the original loss function to better recover the network accuracy.

%% file: sec/motivation.tex
\section{Background and motivation}
\label{sec:motivation}

\textbf{LLM inference. }
LLMs are made of transformer layers. Given an input prompt, each round through this LLM network generates a new token, and the new token is fed into the LLM for generating the token in the next round. Ideally, for the next round, the LLM should need the initial prompt and the answer generated so far as the input context to generate the next token. However, since all the tokens except the last generated token remain the same as the previous round, in order to save on redundant computation, the LLM stores the embeddings for them in KV caches when they are generated for the first time. So in the next round, the LLM simply retrieves the history, state, or embeddings of the previous tokens and processes the last generated token in conjunction with the previous embeddings to generate the next token. The LLM updates the history with the last token and repeats the process until a complete answer is generated. Except for the first round, because the text generation at each step primarily depends on the last generated token, i.e., a single row of input or activation, the text generation phase for a single inference case mainly involves GEMV operations. On the other hand, processing the initial multi-token prompt or text generation for batched inference cases (i.e., many concurrent users) involves many rows of input or activation, necessitating GEMM operations.

\begin{figure}[tb]
\centering
\includegraphics[width=1.0\textwidth]{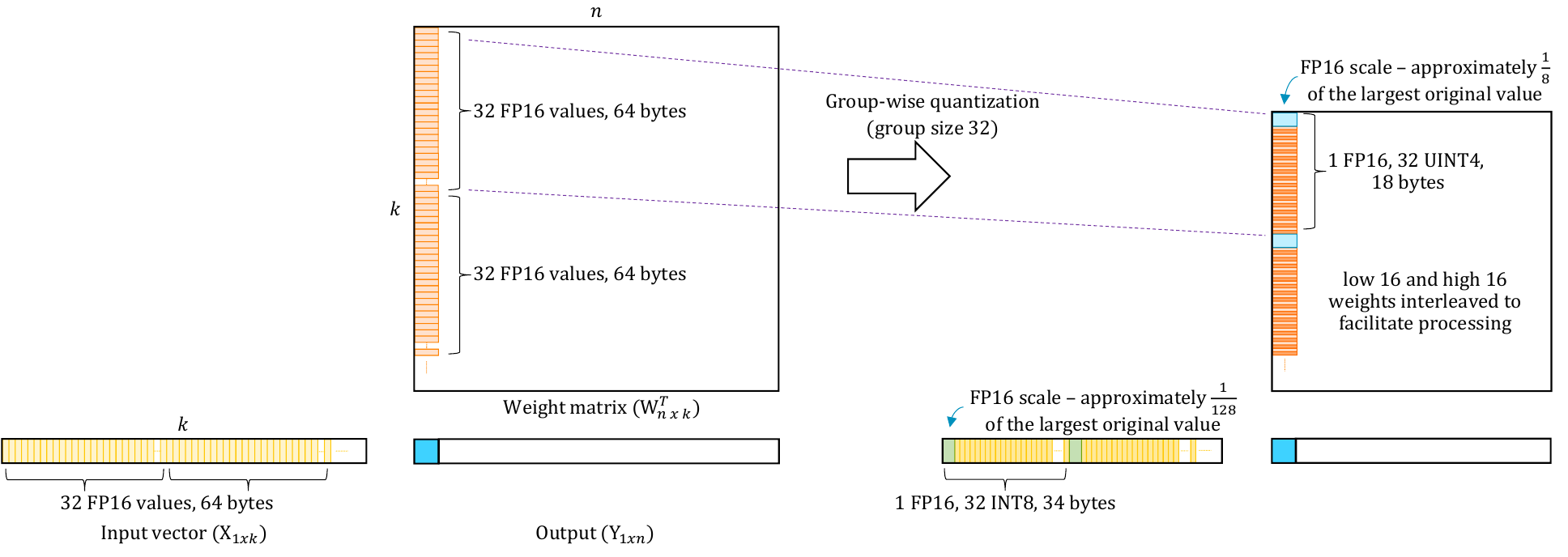}%\vspace{-3mm}
	\caption{Group-wise quantization, in which weights are divided into groups, each with V elements and its own scale factor. We use a group size (V) of $32$ here. Given a weight tensor, a group of $32$ floating-point weights is quantized into $4$-bit integer values using a local scale factor. The next set of $32$ consecutive weights are then quantized to $4$ bits using a different scale factor, and this process is repeated until the entire weight tensor is covered. We use FP16 precision for scale factors.}
\label{fig:group_wise_quantization}
\end{figure}

\textbf{Group-wise quantization. }
For typical operators in LLMs, weight matrices are significantly larger than activation matrices. As a result, compression of the weight matrix is critical to reducing memory and bandwidth consumption, so they are typically quantized to $4$ or fewer bits. Typically, tensor-wise uniform quantization is used for $8$-bit quantization with $256$ distinct quantized values, where a single floating-point scale for the entire tensor can convert the quantized values to actual weights with very low quantization noise. However, quantizing a $16$- or $32$-bit float value to a $4$-bit integer (INT4) or even fewer bits is complex, as an INT4 can only represent $16$ distinct values, compared to the vast range of the FP32. 
%One issue with this tensor-wise quantization approach and others like it is that outlier values can have a disproportionate impact on scaling: the full range of the lower-precision data type isn’t used effectively, which lowers the quantized model’s accuracy.
One issue with this tensor-wide quantization approach is that LLM weight tensors can feature ``outliers'' having much larger magnitude than the other weights; a scale factor chosen to accommodate the outliers results in the remaining weights being represented much less accurately, lowering the quantized model's accuracy.
As a result, when the weight matrix is quantized to $4$ or fewer bits, it is typically quantized using group-wise quantization. Group-wise quantization has a finer granularity than standard tensor-wise or channel-wise quantization~\cite{dai2021vsquant}, allowing it to reduce quantization noise natively while approaching the full-precision (floating point) quality of a foundation model. 
Group-wise quantization quantizes in groups, whereby weights are divided into groups of $32$, $64$, or $256$, as shown in Figure~\ref{fig:group_wise_quantization}. Each group is then quantized individually to mitigate the effect of outliers and increase precision.

For example, the Q4\_0 group quantization format from LLaMA.cpp considers a group size (V) of $32$ and uses an FP16 scale factor to quantize weight values to $4$ bits before interleaving the top and bottom halves of the $32$ $4$-bit weights into $16$ bytes. In the case of activations, size and bandwidth are less important, so they are typically quantized to $8$-bit integer values and the corresponding Q8\_0 format groups and quantizes them to $8$ bits using FP16 scale factors. This weight quantization and interleaving format is chosen to optimize space and bandwidth, as well as to make the decompression process easier during inference, whereas the activation quantization format is chosen to facilitate subsequent integer dot product computation with group-quantized weights, as discussed in the following sections.

\begin{figure}[tb]
\centering
\includegraphics[width=1.0\textwidth]{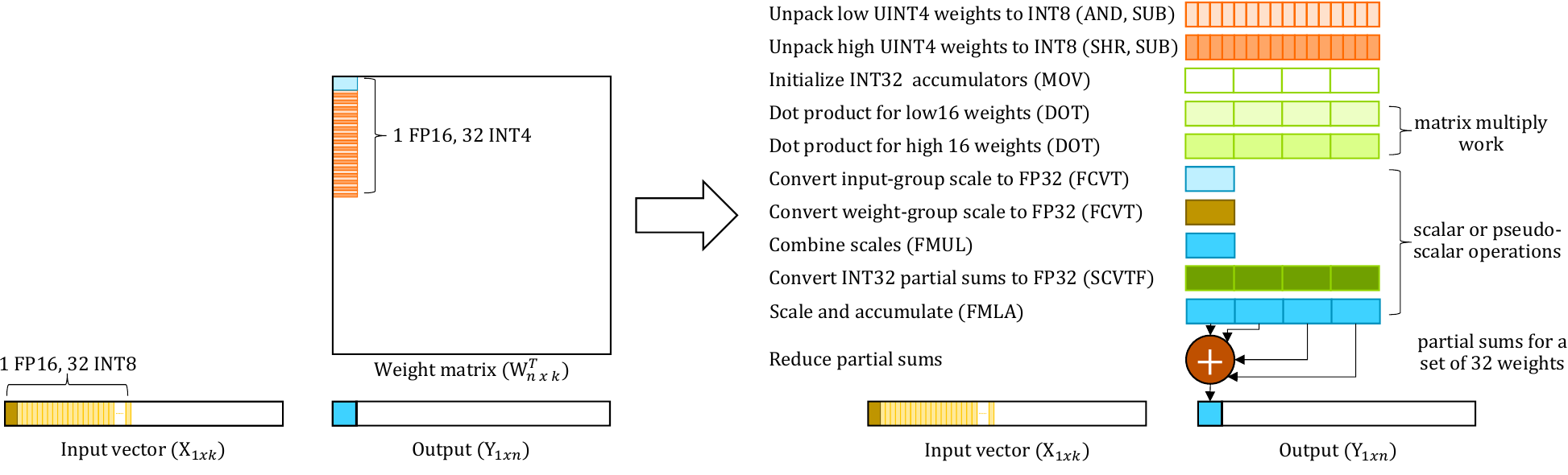}
	\caption{Group processing steps in a reference baseline group-wise quantized dot product kernel.}
\label{fig:ref_dot_product_kernel}
%\vspace{-2mm}
\end{figure}

\textbf{Motivation. }
Because the weights are in a $4$-bit group-quantized format, a matrix-multiply kernel involving group-quantized $4$-bit weights must incur an overhead when dequantizing them, as shown in Figure~\ref{fig:ref_dot_product_kernel}. In particular, they need to be first unpacked and expanded from $4$-bit weights to signed $8$-bit values before computing dot-products between $8$-bit integer weights and activations.
%using integer representations. 
Aside from that, the FP16 scale factors of weights and activations for a group must be expanded to FP32 values before being combined and used to scale a group's integer dot product result (INT32 partial sum), as well as contributing to the final FP32 dot product value for a weight column. This process is repeated until the scaled dot product results of all quantized groups in an entire weight column are combined to obtain the weight column's final dot product value against an activation row.

Due to typically considering a single weight column at a time by the reference kernels of CPU-based LLM inference frameworks, such as LLaMA.cpp, as shown in Figure~\ref{fig:ref_dot_product_kernel}, there is no reuse of activations (input vector) for GEMV kernels or activations and weights for GEMM kernels. This leads to a large number of redundant loads. There is no reuse of the activation's FP16 scale factors or corresponding FP32 converted values. Furthermore, because only one weight column is considered at a time, vector instructions cannot be used to convert FP16 scale factors to FP32 values for a group of weight columns or to combine FP32 scale factors of weights and activations, resulting in a large number of scalar operations. 

Additionally, there are quite a few ``pseudo-scalar'' operations, which operate on a vector of values that is actually one true value split across vector lanes, and are significantly less efficient than ``true'' vector operations. This essentially means that, while fused multiply-accumulate %(FMA)
(dot product) vector operations, such as the ARM dot product operation, \texttt{vdotq\_laneq\_s32}, of the reference matrix multiply kernel operate on vectors of values, they still operate on a single weight column. This, in turn, necessitates additional reduction operations to reduce scaled partial dot products or partial sums from different vector lanes ($4$ vector lanes as shown in Figure~\ref{fig:ref_dot_product_kernel}) in order to obtain the final FP32 dot product result for a weight group and accumulate it to the weight column's overall dot product value.
%to obtain the final FP32 dot product result for a weight group and accumulate it to the overall dot product values for the weight column. 
In summary, many compute instructions do not perform useful matrix multiplication work.

%% file: sec/kernel_optimizations.tex
\section{Arm CPU architecture optimized kernel design for LLMs}
\label{sec:optimized_kernels}
To this end, we present the design of GEMV and GEMM kernels for group-quantized LLMs optimized for various families of Arm CPU architectures. We consider weights and activations to be quantized to $4$ and $8$ bits, respectively, before delving into optimizations for ultra-low-precision weights (e.g., $2$ bits per weight) and non-uniform quantization. Matrix-multiply kernels (GEMV and GEMM) of QKV, output projections, and FFN layers of an LLM operate on $4$-bit weights and $8$-bit activation inputs, primarily carrying out efficient integer computation and generating FP32 outputs. Attention layers perform computations in higher-precision, such as FP16 or FP32. Prior to performing GEMV and GEMM for a LLM layer, FP32 outputs from a previous layer undergo group-wise, dynamic quantization instead of per-tensor, static quantization (i.e., scaling factors computed offline) to produce $8$-bit activation inputs. Dynamic quantization ensures low quantization noise and high accuracy.

\subsection{Optimized GEMV for autoregressive decoding phase}

To increase the reuse of the input activation vector as well as the use of MAC and vector operations, the GEMV kernel considers a series of consecutive weight columns of an LLM weight matrix at a time, as shown in Figure~\ref{fig:simd_aware_weight_packing}. The use of multiple weight columns in the GEMV kernel leads to increased reuse of the quantized activation vector and associated scale factor, as well as fewer load operations. Furthermore, our optimized GEMV kernel uses vector instructions for weight scale factor conversions, enabling it to convert FP16 scale factors of multiple quantized weight groups from different weight columns to FP32 values using a single vector operation. While multiple weight columns improve the reuse of the input vector and the use of vector operations in a GEMV kernel, the overhead from reduction operations and dequantization operations pose a significant challenge to group-quantized GEMVs in achieving good MAC unit utilization and thus a high percentage of useful work. We address the overhead of reduction operations through SIMD-aware weight packing, which interleaves weights from multiple weight columns prior to performing GEMV, and we reduce dequantization overhead by saving signed values directly into group-quantized weights.
%This, combined with SIMD-aware weight packing and fast dequantization of the group-quantized weights, increases the percentage of useful work to $21\%$ for 4 weight columns. Since you are rearranging the weights for these optimizations, it can be done offline, so that cost goes away. Furthermore, by saving signed values directly for the top nibbles, you can get rid of subtraction operations during the dequant path. This further improves the MAC efficiency to $31\%$, which is an $85\%$ speedup over the original code.

% Half the operations in original code are scalar or “pseudo-scalar” – operating on a vector of values which is really one true value split across lanes.
% This technique reduces the number of reduction operations (sum across lanes) needed.
% Still less efficient than ”true” vector operations.

% Using true vector operations should improve performance by around 60\%.
% Runtime of 50\% (already vectorized) + (50\%/4 = 12.5\%) = 62.5\%.
% $62.5\%$ runtime = 1.6x performance.

% => Vector lanes need to accumulate different results rather than multiple parts of the same result.
% => Compute more than one result at once – for non-batched case this must be different output points.

\subsubsection{SIMD-aware weight packing}

\begin{figure}[tb]
\centering
\includegraphics[width=1.0\textwidth]{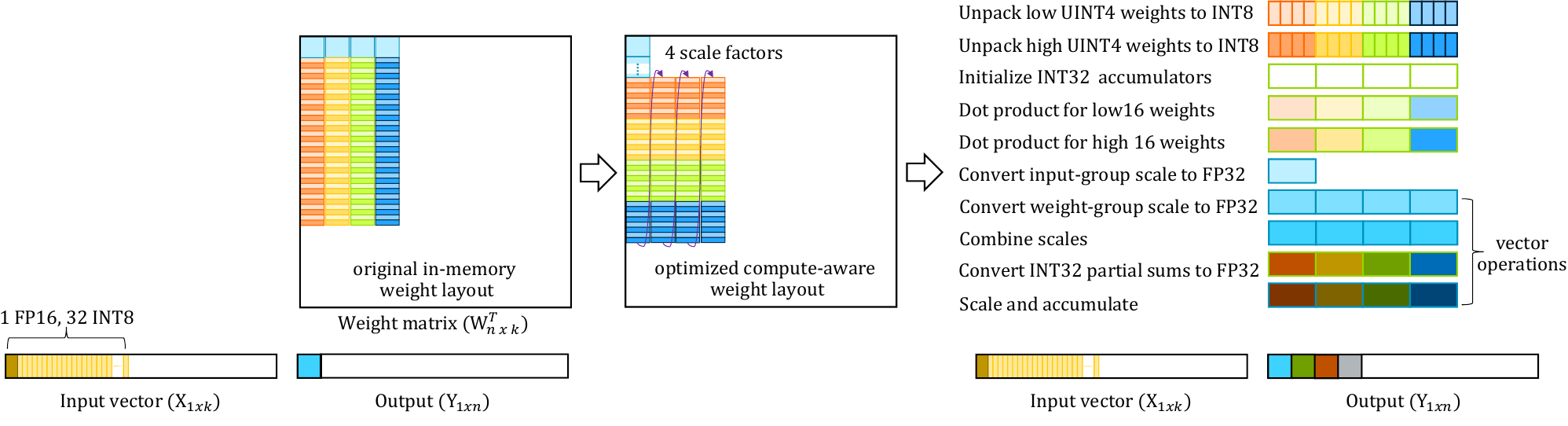}
	\caption{SIMD-aware weight reorder to minimize scalar operations in GEMV and GEMM kernels.}
\label{fig:simd_aware_weight_packing}
\vspace{-2mm}
\end{figure}

Before decompressing quantized operands and performing the necessary arithmetic operations of a matrix multiply kernel, the operands must be loaded from memory into the register file. Naive loading of $4$-bit consecutive weight elements from a quantized group of a single output channel requires the 
fused multiply-accumulate %(FMA) 
(dot product) instruction \texttt{vdotq\_laneq\_s32} to perform additional reduction operations. Reduction operations must be performed on partial dot products from different vector lanes to obtain the final dot product result for a quantized weight group, as discussed in Section~\ref{sec:motivation} and Figure~\ref{fig:ref_dot_product_kernel}.
This can be avoided if different vector lanes of the \texttt{vdotq\_laneq\_s32} instruction operate on weight elements from different output channels. This ensures that the accumulators for the \texttt{vdotq\_laneq\_s32} instruction's various vector lanes can accumulate results from different output channels rather than multiple parts of the same output channel. This in turn necessitates a strided weight layout for each vector lane in computation, as illustrated in Figure~\ref{fig:simd_aware_weight_packing} (middle).
A naive weight loading scheme would necessitate loading the corresponding quantized group from multiple output channels, incurring additional overhead from pointer arithmetic operations and address calculation for each output channel. Furthermore, non-contiguous access patterns of quantized groups across channels in memory prevent achieving the best possible DRAM bandwidth.

We solved this problem by storing the group-quantized weights from consecutive output channels in memory in the same order that they are used during computation. Instead of permuting weights each time, we reformat them beforehand to match the compute order and store them in memory in reordered format. This reordering has no runtime overhead because it is performed permanently on weights before they are used for inference.
In order to perform compute-aware weight reordering, we first store the scale factor of the quantized groups of several consecutive output channels, then reorder and store the corresponding quantized elements from them. For example, when reordering quantized weight groups with a group size of $32$ from four output channels, the first four bytes of the four quantized groups are stored one after the other from four output channels, followed by the next four bytes of the quantized groups. This process is repeated until all of the bytes from these groups are stored in the reordered format. After the first four groups from these four output channels are reordered, the next four groups from these output channels are considered and reordered in the same manner. This process continues until all of the quantized groups from these output channels are reordered before proceeding to the next set of four output weight channels.
%Such that each vector lane eventually find the required data for int8 tenor core computation.

This reordering is space-neutral, with the same data stored in a different order. In addition to improving the locality of reference and bandwidth of memory transactions, as well as lowering pointer arithmetic overhead, it improves the alignment properties of quantized groups in memory. For a group size of $32$, $4$-bit quantized weights, and the FP16 scale factor, this ensures that no more $18$-byte structures are stored in memory. Instead, better-aligned, reordered weight groups from consecutive channels (e.g., an $8$-way grouped structure of $144$ bytes for weight groups from eight channels) are stored. It also simplifies scale factor handling by eliminating the need to assemble vectors from multiple locations.
% Instead of permuting weights each time, store in memory in blocked format instead.
% Space neutral – same data in a different order.
% Improved alignment characteristics (no more 18-byte structures).
% Scale factor handling easier (don’t need to assemble vector from multiple locations)
% Could go full “structure of arrays”; we just went for ”array of more useful structures”.

\subsubsection{Fast dequantization}

Because the current Arm CPU architectures do not support multiplication of $4$-bit and $8$-bit values, or $4$-bit and $4$-bit values, for quantized layers, dequantizing operands to $8$-bit within the GEMV and GEMM kernels is required prior to performing matrix multiply computation. Our proposed matrix multiply kernels for Arm CPUs fuse dequantization kernels with matrix multiplication kernels to avoid writing dequantized values to DRAM. In the case of $4$-bit weight quantization, unsigned $4$-bit integers should be unpacked into unsigned $8$-bit integers before being converted to signed $8$-bit integers within a matrix multiply kernel.

For $4$-bit group quantization formats, typically the top and bottom halves of the $4$-bit weights of a group are interleaved in memory. 
%As illustrated in Figure~\ref{fig:4b_orig_weight_packing}, 
For a group size of $32$, the $4$-bit weights $w_0$, $w_1$, $w_2$, $w_3$, ..., $w_{31}$ are reordered into $w_0$, $w_{16}$, $w_1$, $w_{17}$, ..., $w_{15}$, $w_{31}$ sequence and stored in $16$ bytes. Furthermore, to avoid sign extension issues, signed $4$-bit values are stored as unsigned after adding an $8$-bit bias value~\cite{lin2024awq}. The same ordering within each group is preserved when group-quantized weights from multiple channels are interleaved to match the compute order, as described in the previous section. This weight packing format was specifically chosen to efficiently unpack them into unsigned $8$-bit values using a few SIMD operations (bitwise AND and shift operations), and then subtract $8$ to restore true signed values and sign bits of the $4$-bit nibbles during the dequantization path.
%, as shown in the Figure. 
Although simple, this method adds a significant overhead to the dequantization process.

\begin{figure}[tb]
\centering
\includegraphics[width=0.8\textwidth]{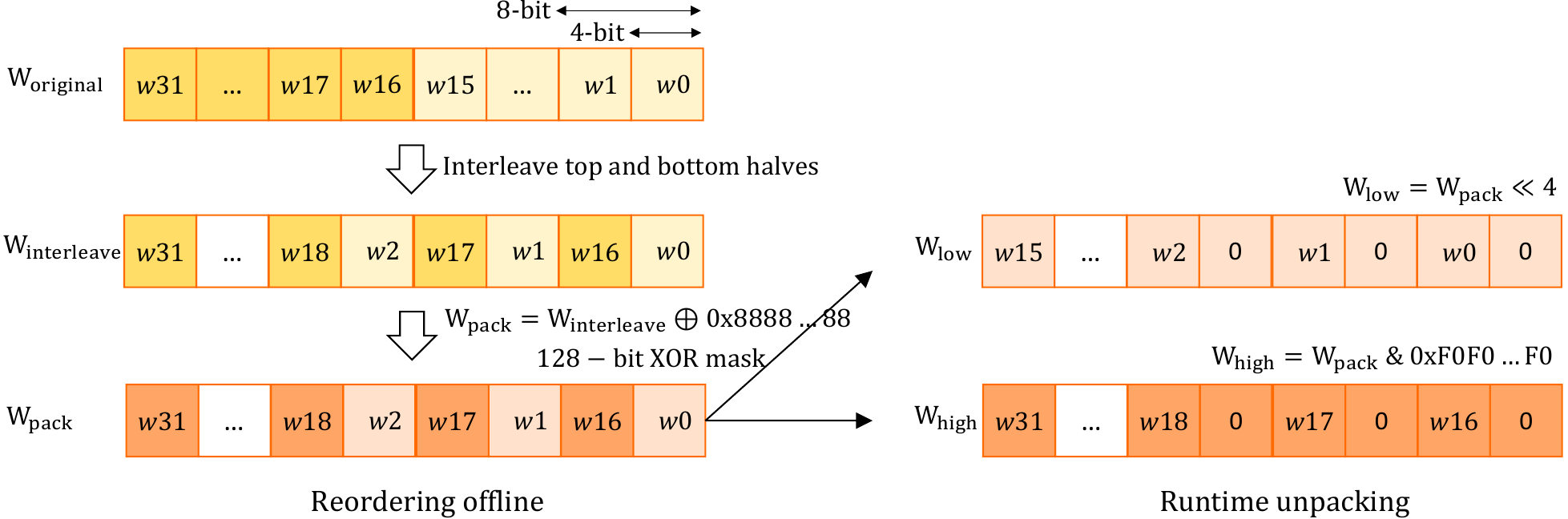}
	\caption{Fast decompression path for unpacking 4-bit nibbles into signed 8-bit weights in GEMV and GEMM kernels.}
\label{fig:fast_decompression}
\vspace{-2.25mm}
\end{figure}

Instead, we employ a more efficient approach for storing signed values directly, significantly reducing the dequantization overhead of converting them to signed $8$-bit values during matrix multiplication operations, as illustrated in Figure~\ref{fig:fast_decompression}. This is accomplished by toggling the most significant bit (MSB) of each nibble stored in the byte (stored as unsigned after adding an $8$-bit bias value, as previously stated) during the weight reordering stage. After this operation, the two nibbles represent the four most significant bits of the original signed nibble values after being converted to signed $8$-bit values and multiplied by $16$. Therefore, at runtime, we can simply use an arithmetic shift left to get the low nibble (bits $0$-$3$) and a 0xF0 mask to get the high nibble (bits $4$-$7$) of a byte, because the arithmetic shift left fills the empty bits with zeros. Because these weight values are scaled up by $16$, we must scale down the subsequent partial dot product value that uses them to obtain the correct result. 
%We accommodate it in the following operation, which converts the signed integer partial dot product value to a floating-point value, to fix up the values. The integer-to-floating-point convert instruction is required in the group-quantized matrix multiply kernels to scale the integer dot product with the group's scale factor before adding it to the floating-point master accumulator.
We accomplish this by dividing the partial product by $16$ using either an implicit shift right (as part of the required integer-to-floating-point convert instruction in group-quantized matrix multiply kernels to update the floating-point master accumulator) or an explicit floating-point exponent adjustment operation. It is significantly less expensive than having to subtract $8$ from each vector of weights and saves eight subtraction operations, requiring only one scale operation on the accumulator at most.
%\vspace{-1mm}

\subsection{Optimized GEMM for prompt phase (time-to-first-token)}

While GEMV operations make up the majority of the computations in the decode stage for a single inference case (i.e., a request from a single user), GEMM operations dominate the prompt phase (prefill stage) for both single and batched inference cases. Furthermore, for batched inference cases, GEMM consumes the majority of the computation for the decode stage as well.
 
We apply the same optimizations to developing Arm CPU architecture optimized GEMM kernels as we do for GEMV kernels. In particular, in order to increase compute throughput further, maximize the use of vector operations, and avoid pseudo-scalar operations, we make use of the matrix-matrix multiply-accumulate (MMLA) instruction. The MMLA instruction can perform twice as many MAC operations when compared to an equivalent SIMD dot product instruction (for example, $128$-bit DOT) used in the GEMV kernel. An $128$-bit MMLA instruction performs double operations by processing multiple rows of activations at once. It multiplies a $2$x$8$ matrix of $8$-bit integer values in the first source vector by an $8$x$2$ matrix of $8$-bit integer values in the second source vector to produce a $2$x$2$ matrix of $32$-bit integer product values.

However, this requires reordering activations from multiple input rows to correspond to the weight reordering and value ordering required by the MMLA operation. Because each matrix multiply kernel is always preceded by a dynamic re-quantization kernel for FP32 activations, the reorder kernel for activations can be fused into it with minimal latency overhead. Furthermore, a larger number of input activation rows and output weight channels creates high pressure on vector register files for storing partial dot products, causing large GEMM kernels to be register-bound on Arm CPUs due to the nature of the output stationary dataflow. As a result, the size of the vector register file for an Arm processor architecture type influences the number of concurrently processed rows and columns and the design of a GEMM kernel. In particular, we design three types of group-quantized GEMV and GEMM kernels based on specifications such as the availability of SDOT or MMLA instructions, weight channel interleaving patterns, SIMD vector widths, and register file sizes of different available Arm CPU types.

\subsection{Turning intrinsics into assembly}

While we can generally rely on the compiler and use regular intrinsics for the group-quantized GEMV and GEMM kernels, we always find that the compiler does not generate fully optimized code, especially when there is high register pressure while running GEMMs during the prefill stage for single inferences and the prefill and decode stages for batched inferences. As a result, we convert intrinsics into assembly code to maximize the use of available vector registers and MAC units, avoid spilling of register values to memory and associated restore code, and improve instruction-level parallelism, leading to improved compute efficiency.

%Are these hand-written? Can we not rely on the compiler and write regular intrinsics instead?
%QServe exploits register-level parallelism to significantly reduce the number of required logical operations in UINT4 to UINT8 weight unpacking.
%Auto vectorization leading sub optimal code, hence we proposed assembly code

%\subsection{Kernel fusion}
%We also extensively apply kernel fusion to optimize on-device LLM inference. For layer normalization, we fuse all operators (e.g. multiplication, division and square root) into a single kernel. For attention layers, we fuse QKV projections into a single kernel, and also perform on-the-fly positional embedding calculation. We also preallocate KV caches and perform cache updates within the attention kernel. Kernel fusion is particularly useful for models with inefficient forward pass implementations, such as Falcon (Penedo et al., 2023) and StarCoder (Li et al., 2023c). Notably, the computation time for each FP16 kernel is in the order of 0.01ms on the 4090 GPU, comparable to the GPU kernel launch overhead. Hence, reducing number of kernel calls through kernel fusion leads to direct speedups.

%% file: sec/nonuniform_quant.tex
\section{Group-wise non-uniform codebook-based quantization}
\label{sec:nonuniform_quant}

Uniform quantization divides the range of weight values into equal intervals and assigns a quantization level to each interval. It distributes quantized values uniformly and equidistantly. As a result, despite being commonly used in conjunction with group-wise quantization for LLMs, it is not very flexible in matching the non-uniform patterns typically found in LLM weight distributions~\cite{kim2024sqllm}, 
%as illustrated in Figure \ref{fig:llama2_7b_weight_distribution},
resulting in suboptimal accuracy, especially for low-precision LLM quantization. Non-uniform quantization with a codebook allows for a more flexible allocation of high-precision weight values. Given a weight distribution, non-uniform codebook-based quantization can identify $k$ centroids that best represent the weight values and map weights to them. For example, when quantizing a weight distribution to $4$-bits, state-of-the-art codebook-based quantization techniques aim to determine the $16$ centroid values that best represent the values. Each high-precision weight can then be represented by the $4$-bit index of a centroid in the codebook instead of its original bit-width. In addition, non-uniform codebook-based 
quantization requires storing the codebook itself and incurring associative overhead.
While there are a few recent non-uniform codebook-based quantization techniques for LLMs in the literature~\cite{kim2024sqllm, vanbaalen2024gptvq, chee2023quip, tseng2024quip}, they either do not exhibit good runtime for both phases (prompt processing and autoregressive decoding) of LLM inference under ultra-low-bit precision scenarios or do not extend well to achieving extreme degrees of compression, as discussed below.
\subsection{Challenges of prior non-uniform quantization techniques}
%We conduct an extensive study of ultra-low-bit precision non-uniform codebook-based quantization techniques in the literature to better understand their potential and limitations.
\textbf{SqueezeLLM~\citep{kim2024sqllm}}. In order to be more sensitive to the importance of the weights, SqueezeLLM quantization first clusters the weights using a weighted k-means clustering algorithm where the centroids of the cluster (codebook) are chosen to be close to the sensitive weights. In other words, rather than scaling high-precision weights group-wise into the range provided by a given number of bits, SqueezeLLM uses weighted k-means clustering on all weights in a tensor row, mapping weights to codebooks with the number of codebooks determined by the bit per weight a quantization scheme wishes to spend. However, the improvement in representation of the weight distribution by the SqueezeLLM quantization comes at the cost of loading the codebook for each row of a weight matrix along with the index assignments for the weights. This, combined with FP16 values for the per-row codebook entries, results in inefficient floating-point computation and significantly slows down LLM inference, as observed in our evaluations. Furthermore, before using the SqueezeLLM quantization, the codebooks for each layer of an LLM must be determined; the same codebook entries will not work for an unseen LLM. In addition, SqueezeLLM does not support low-precision quantization below $4$-bits, which is required to fit large-scale LLMs to resource-constrained devices and is thus the primary focus of our proposed method.

\textbf{GPTVQ~\citep{vanbaalen2024gptvq}}. Recent work GPTVQ extends the potential of non-uniform quantization for higher levels of compression (for example, $2$-bit and $3$-bit quantization) and outperforms its uniform counterpart. Notably, it makes use of vector quantization, which involves quantizing multiple weights together and mapping them to a single centroid in a codebook rather than representing each quantized value with a centroid in the codebook, resulting in a more versatile quantization grid across multiple dimensions. It also performs codebook quantization to $8$-bits and shares the codebook across multiple rows/columns of an LLM layer. While it demonstrates the potential of codebook-based quantization for ultra-low-precision quantization, the accuracy of the resulting LLMs suffers noticeably (for example, for $2$-bit quantization).

\textbf{QuIP\#~\citep{tseng2024quip}}. State-of-the-art $2$-bit quantization technique QuIP\# improves upon previous work by leveraging lattice codebooks for vector quantization and incoherence processing for superior outlier suppression. E8 lattice codebooks encode the magnitude of quantized values in a group of eight. Besides, QuIP\# forces an even number of positive (or negative) signs of quantized values in a group of eight. This enables the use of seven bits to record the sign of eight quantized values. While the combination of these optimizations allows QuIP\# to achieve good LLM quality in extreme compression regimes (for example, $2$ bits per weight), the resultant complexity of the decompression path in converting $2$-bit quantized weight values to $8$-bit significantly slows down LLM inference, as found in our evaluations.

Our proposal addresses limitations in prior non-uniform quantization techniques and attempts to fill the void by not only ensuring faster throughput but also better quality than the current state-of-the-art under extreme compression scenarios (e.g., $2$-bit quantization) for LLMs.

\subsection{The group-wise codebook-based quantization method}

\begin{figure}[tb]
\centering
\includegraphics[width=0.8\textwidth]{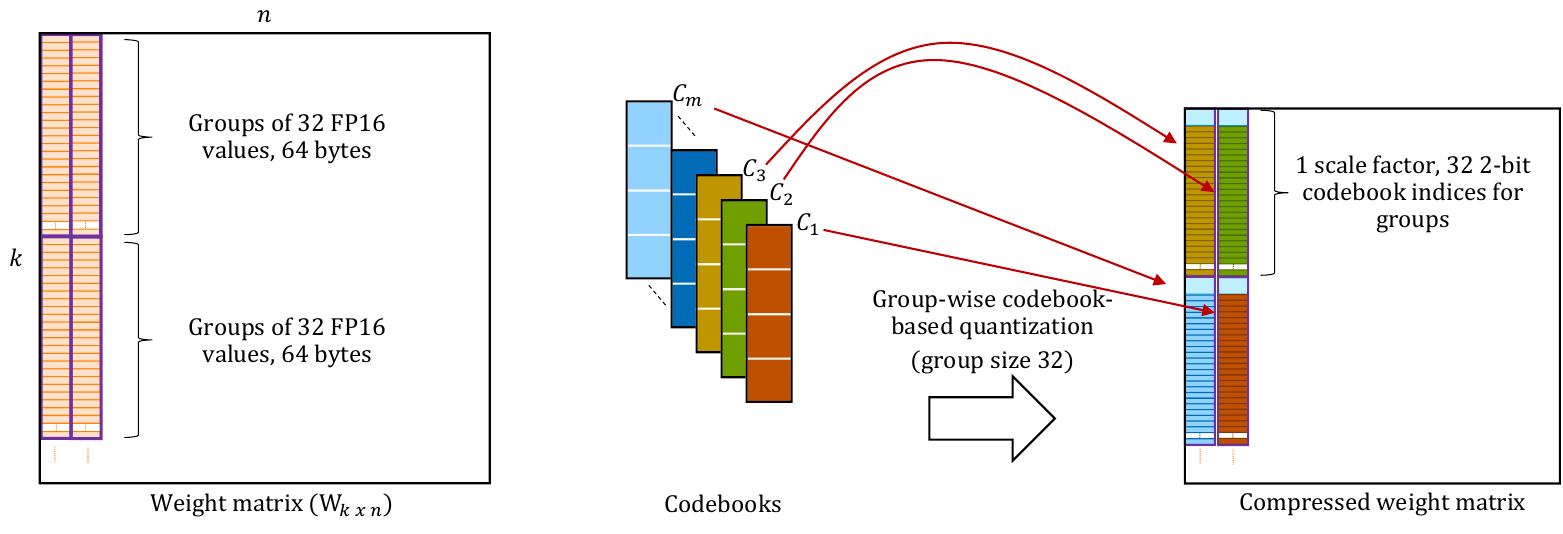}
	\caption{Fine-grained assignment of codebooks to various groups in group-wise codebook-based quantization. Each group finds the closest codebook of the $C$ codebooks ($C_1$, $C_2$, ..., $C_m$) that best represents its values and quantizes its high-precision values to the codebook centroids using $2$-bit indices.}
    %Groups of weights are represented by a sum of codes selected from codebooks by corresponding indices}
\label{fig:groupwise_cb_quantization}
\end{figure}

Our innovation is motivated by the following observations and insights: For LLM weight matrices, which are commonly quantized group-wise, there may be some variations in the shape of the Gaussian distribution of values between groups. However, after being scaled by the group-wise scale factors, the Gaussian distributions of various groups with different shapes should be clustered into a small set of shapes, each of which can be represented with its own codebook.

Motivated by these insights, we apply a group-wise structure to divide the high bit-width floating point weights into groups and scale each group separately first, using its own scale factor. The scale factor is chosen so that the ranges of values after scaling can be represented by codebook's bit-width. For example, if the required bit-width of centroid values in a codebook is a signed $8$-bit integer, the scale factor for a group scales the group's weights to the $-128$ to $127$ range. We then use a two-phase clustering algorithm (constraint-satisfaction-guided clustering algorithm) to cluster the scaled group of weight values into a small number of codebooks, each with a few centroid values. In phase $1$, similar weight groups or groups of scaled weight values with similar Gaussian distributions are divided into $C$ clusters by converting each group of scaled weight values to a probability distribution. This is accomplished by finding the histogram of the scaled weight values and normalizing it, converting the discrete distribution of intensities into a discrete distribution of probabilities, and then applying k-means clustering to these probability distributions, which now represent the different groups, to cluster the similar groups. Phase $2$ then applies k-means clustering analysis to each clustered group of values (created in phase $1$) to identify a few centroid values that best represent the probability distribution of weight values within each and repeats it for $C$ clustered groups of values to create $C$ codebooks to find the different non-uniform weight distributions present in the high-precision weights. %The clustering of similar groups into the same codebook aids a group in locating the closest codebook that best represents its values
The clustering of similar groups into the same codebook aids a weight group later during post-training quantization in locating the closest codebook of the $C$ codebooks that best represents its values, as shown in Figure~\ref{fig:groupwise_cb_quantization}, while the small number of centroid values for each codebook ensures that high-precision centroid values, such as four $8$-bit signed centroids in a codebook, can be encoded using a lower bit-width index ($2$-bit indices here) in the codebook.

For extreme compression cases, we observe that quantization bit-width and associated number of 
distinct values or quantization bins (e.g., $16$ distinct values for $4$-bit quantization) have the greatest impact on LLM accuracy. Changes in the bit-precision of scale factors, on the other hand, have less of an impact on LLM accuracy. The various codebooks in group-wise codebook-based quantization essentially help in adapting a group to choose a subset of relatively higher-precision data type while using low bit-width indices ($4$ most important quantization bins of a distribution for a group here using $2$-bit indices). This results in closely following the distribution of quantization bins of a higher-precision data type and bridging the accuracy gap with it.

Furthermore, our codebook-based quantization scheme keeps the decompression path in converting 
low-bit codebook indices to high-precision centroid values simple when compared to prior schemes, resulting in improved throughput.
Given high-precision weight values of a LLM, Algorithm~\ref{alg:q2nl_quantization} describes how to generate a predetermined number of codebooks and subsequently use them in post-training quantization of the LLM, while Algorithm~\ref{alg:q2nl_inference} describes fast inference with a group-wise codebook-based quantized layer.

\subsection{Example group-wise codebook-based quantization for 2-bit width}
\begin{algorithm}[t]
%\caption{Q2\_NL codebook-based quantization: Quantize W given the number of codebooks c, the number of centroids k, the group size l, and the super-group size s}\label{alg:q2nl_quantization}
\caption{Fine-grained codebook-based quantization: codebooks creation and post-training quantization }\label{alg:q2nl_quantization}
\begin{algorithmic}[1]

% \Require $n \geq 0$
% \Ensure $y = x^n$
% \State $y \gets 1$
% \State $X \gets x$
% \State $N \gets n$
% \While{$N \neq 0$}
% \If{$N$ is even}
%     \State $X \gets X \times X$
%     \State $N \gets \frac{N}{2}$  \Comment{This is a comment}
% \ElsIf{$N$ is odd}
%     \State $y \gets y \times X$
%     \State $N \gets N - 1$
% \EndIf
% \EndWhile
%Q2_NL Non-uniform codebook-based quantization: 
\Require High-precision weights $W$, divided into groups of size $g$
%\\
%\State Step 1: Scale each group of high-precision weights to the -128 to 127 range by using FP16 scale factors.
% \State Step 2: Cluster the similar weight groups created in Step 1 (groups of scaled weight values with similar gaussian distributions) into C clusters by performing the following steps:

% -	Convert each group of scaled weight values to a probability distribution. One method is to scale each group of values to a smaller range (for example, [-m, m] range, where m = 8), then find the histogram of the scaled values and normalize it, converting the discrete distribution of intensities into a discrete distribution of probabilities.

% -	Apply clustering analysis (for example, k-means clustering) to these probability distributions, which now represent the different groups, to cluster the similar groups.
\For{\texttt{each group of high-precision input weight values}}\algorithmiccomment{Codebooks creation}
    \State \texttt{scaled input $\gets$ input weight values $\div$ FP16 scale factor}
    \State \texttt{normalized per-group distribution $\gets$ Histogram (scaled input)}
    %\State \texttt{clustered group of values $\gets$ k-means clustering (normalized distribution)}
    \State \texttt{$D$ $\gets$ $D$ $\cup$ normalized per-group distribution}
\EndFor
\State \texttt{clustered group of values $\gets$ k-means clustering ($D$)}
%\State Step 3: Apply clustering analysis now to each clustered group of values (created in Step 2) to identify a few centroid values that best represent the probability distribution of weight values within each. Repeat Step 3 for C clustered groups of values to create C codebooks to find the different non-uniform weight distributions present in the high-precision weights W.

\For{\texttt{each clustered group of values}}
    \State \texttt{centroid values in $c_i$ $\gets$ k-means clustering to create codebook $c_i$}
    \State \texttt{codebooks $C$ $\gets$ $C$ $\cup$ $c_i$}
\EndFor

% \State Step 4: Use C codebooks created in Step 3 to perform post-training quantization of high-precision input weight values by performing the following steps:

% -	group of scaled input weight values ← group of high-precision input weight values/FP16 scale factor 

% -	codebook index, centroid indices for a group ← For a group of scaled input weight values, choose a codebook (of the C codebooks) that best matches their distribution with the lowest reconstruction MSE (mean square error), and map weights to the centroid values in the chosen codebook. Output encoded data representative of the selected codebook and mapped centroid values.

% Repeat for each group of high-precision input weight values.
%\\
\For{\texttt{each group of high-precision input weight values}}\algorithmiccomment{PTQ using codebooks}
    \State \texttt{scaled input $\gets$ input weight values $\div$ FP16 scale factor}
    \State \texttt{$c_j$ $\gets$ codebook $c$ $\in$ $C$ that best matches scaled input's distribution}
    %\State \texttt{centroid indices in $c_j$ $\gets$ map scaled input to the centroid values in $c_j$}
    \State \texttt{centroid indices in $c_j$ $\gets$ map scaled input to centroid values in $c_j$}
\EndFor
%\\
\Ensure $C$ codebooks, $FP16$ scale factor, codebook index, and corresponding centroid indices for each group of high-precision input weight values

% \State Step1: cluster the similar groups (groups of similar gaussian distribution)
% 	[a, b, c, … ], [m, n, o, …], [x, y, z, …], [d, e, f, …], [i, j, k, …], [r, s, t, …]
 
% 		- scale the groups to [-8, 8] range, sort the values, apply histogram and 				turn the groups into probability distribution
  
% 		- Apply clustering on group-size vectors to cluster the similar groups
% 			(group-size = 16 here) 

% \State Step2: Cluster individual clustered groups (created in step 1) to represent the same 			distribution clusters with few centroid values.

% 	[a, b, c, …, d, e, f, …], [m, n, o, …, r, s, t, …], [x, y, z, …], [i, j, k, …], 

\end{algorithmic}
\end{algorithm}

\begin{algorithm}[t]
\caption{Fine-grained codebook-based quantization: Inference}\label{alg:q2nl_inference}
\begin{algorithmic}[1]
% input Weight W ∈ R
% m×n, hessians H ∈ R
% n×n, g-dim. kbit codebook C
% \Require $n \geq 0$
% \Ensure $y = x^n$
% \State $y \gets 1$
% \State $X \gets x$
% \State $N \gets n$
% \While{$N \neq 0$}
% \If{$N$ is even}
%     \State $X \gets X \times X$
%     \State $N \gets \frac{N}{2}$  \Comment{This is a comment}
% \ElsIf{$N$ is odd}
%     \State $y \gets y \times X$
%     \State $N \gets N - 1$
% \EndIf
% \EndWhile
% \State 4-b index = 2-b | clusterid << 2
%Q2_NL Decompression and inference 
\Require $C$ codebooks, $FP16$ scale factor, codebook index, and corresponding centroid indices for each group of quantized weight values from Algorithm $1$
%\\
\For{\texttt{each group of quantized weight values}}
    \State \texttt{centroid values $\gets$ $C$[codebook index][centroid indices]}
    \State \texttt{decompressed values $\gets$ FP16 scale factor $\times$ centroid values}
\EndFor
%\\
\Ensure Decompressed model weights $W$, divided into groups of size $g$
\end{algorithmic}
\end{algorithm}
We apply our codebook-based quantization technique to compress LLMs to about $2$, $3$, or $4$ bits per weight. Our $2$-bit codebook-based quantization scheme uses a small number of codebooks, such as four, eight, or sixteen, depending on the required compression size and accuracy. As demonstrated in Algorithm~\ref{alg:q2nl_quantization}, they are discovered by first applying a group-wise structure and scale factors to LLM weight matrices, followed by dividing and clustering similar groups (groups with similar distributions) into a small number. The small number of clustered groups then use a clustering algorithm individually to cluster values in them into an equal number of codebooks, each with four centroids. Later, during post-training quantization of a group-wise structured LLM, the various groups choose one of the codebooks that best matches their distribution with the lowest reconstruction MSE (mean square error). As a result, each group typically requires two to four bits to encode the selected codebook's index, as well as two bits for each of its elements to encode one of the codebook's four centroids.

For example, for some $2$-bit linear layers in LLMs, our $2$-bit codebook-based quantization technique has four codebooks, each with four signed $8$-bit integer centroids. It has a group size of $256$, divided into $16$ sub-groups of $16$ $2$-bit quantized elements, each with their own local scale factor. Each sub-group also has a $2$-bit codebook index, which is used to index into one of the four codebooks and extract centroid values corresponding to the $2$-bit index elements. It also has an FP16 superblock scale factor for the $256$-wide group. Our codebook-based quantization scheme is also extended for other bit-widths, such as $3$- and $4$-bit quantization, by dividing similar groups into a few clusters (for example, four, eight, or sixteen) and then encoding each clustered group with an $8$- (for $3$-bit index) or 16-entry (for $4$-bit index) codebook.

%% file: sec/evaluation.tex
\section{Experiments}
\label{sec:evaluation}

%We assess the effectiveness and precision of our implementation by comparing it with the Python bindings for llama.cpp (Gerganov, 2023), and by presenting the perplexity values on the standard wikitext2 dataset (Merity et al., 2016). For this preliminary version, we have executed our generator on the AVX2 instruction set. However, the instructions that we use have equivalents on all SIMD vector architectures.

\subsection{Evaluation setup}

We compare our optimized kernels and codebook-based quantization method to LLaMA.cpp, both in terms of inference throughput (tokens generated / second) as well as in terms of accuracy of the resulting models, measured in terms of perplexity (PPL). 
%We use models generated using the GPTQ quantization method (Frantar et al., 2022), whereas llama.cpp uses pre-generated models using their custom q4 0 quantization format. For performance measurements we use an AMD EPYC 7742 64-Core processor running 64 threads. We compile our code using gcc 9.4.0 with -O3, -mavx, -mavx2, -mfma, -march=native, -ffast-math, -ftree-vectorize flags, and we parallelize using OpenMP 4.5. We compile llama.cpp using OpenBLAS.
LLaMA.cpp lowers the entire computation graph to C++ to minimize overhead on CPUs. We use a prompt sequence length of $128$ and an output token generation length of $128$, and FlashAttention is enabled for all throughput measurement experiments.

\subsection{Inference throughput for 4-bit uniform quantization}
Table~\ref{tab:q4_0_vs_q4_0_8_8_perfcomparison_grav3} compares the runtime performance of our optimized $4$-bit group-wise quantized kernel (Q4\_0\_8\_8) to that of LLaMA.cpp’s $4$-bit kernel (Q4\_0) on Graviton3 processors with $64$ Arm Neoverse V1 CPU cores for different batch sizes (number of users). For the LLaMA-3 8B model, Q4\_0\_8\_8 improves inference throughput (tokens per second) by $3$-$3.2\times$ during the prefill stage and by up to $2\times$ during the autoregressive decoding or token generation stage. The token generation phase for a bath size of one at high core (thread) counts is memory bound, so our optimized GEMV kernels for it, while offering a significant speedup\footnote{For a batch size of one, autoregressive decoding is compute bound for low thread counts, so our optimized $4$-bit GEMV kernels achieve about a $2\times$ improvement in throughput at lower thread counts of up to $16$ in a $64$-core Graviton3 processor.} for a smaller number of cores, cannot provide a tangible improvement at $64$ cores over LLaMA.cpp's reference Q4\_0 kernel. In the case of the LLaMA.cpp FP16 implementation, the prefill rate and token generation throughput increase from $123.5$ tokens/s to $136.2$ tokens/s and $16.9$ tokens/s to $106.4$ tokens/s, respectively, as batch size increases from $1$ to $32$. While the LLaMA.cpp's reference Q4\_0 kernels can offer some improvement in throughput over their FP16 implementation, our $4$-bit optimized kernels improved end-to-end throughput significantly over FP16 and Q4\_0, as shown in Table~\ref{tab:q4_0_vs_q4_0_8_8_perfcomparison_grav3}.

%\begin{table*}[!t]
\begin{table}[t]
   \caption{Comparison of the prefill rate and token generation throughput (tokens/sec) of the LLaMA-3 8B and LLaMA-3.2 3B models for the reference llama.cpp 4-bit uniform quantization kernel (Q4\_0) and our corresponding optimized kernel (Q4\_0\_8\_8) on Arm Graviton3 CPUs (64 cores).}
\label{tab:q4_0_vs_q4_0_8_8_perfcomparison_grav3}
%\vskip 0.15in
\begin{center}
\begin{small}
%\begin{sc}
%\begin{tabular}{|c|c|c|c|c|c|}
\scalebox{0.9}{
\begin{tabular}{crrcrrcrrcrr}
\toprule
& \multicolumn{5}{c}{LLaMA-3 8B} & & \multicolumn{5}{c}{LLaMA-3.2 3B}\\
\cline{2-6}
\cline{8-12}
\\
Batch & \multicolumn{2}{c}{Q4\_0 (llama.cpp)} & & \multicolumn{2}{c}{Q4\_0\_8\_8 (Optimized)} & & \multicolumn{2}{c}{Q4\_0 (llama.cpp)} & & \multicolumn{2}{c}{Q4\_0\_8\_8 (Optimized)} \\
Size & Prefill & Token Gen. & & Prefill & Token Gen. & & Prefill & Token Gen. & & Prefill & Token Gen.\\
\cline{2-3}
\cline{5-6}
\cline{8-9}
\cline{11-12}
%\cmidrule{2-3}
\\
1 & 190.4 & 46.0 & & 570.9 & 46.8 & & 409.8 & 84.3 & & 1137.2 & 87.7 \\
4 & 204.2 & 112.1 & & 650.2 & 149.8 & & 473.0 & 195.7 & & 1472.5 & 270.6\\
8 & 223.8 & 139.1 & & 683.2 & 199.5 & & 538.7 & 268.3 & & 1588.8 & 371.4\\
16 & 222.8 & 157.5 & & 678.5 & 315.8 & & 538.8 & 313.1 & & 1579.3 & 535.4\\
32 & 222.2 & 166.8 & & 665.0 & 342.9 & & 533.7 & 339.0 & & 1550.0 & 585.0\\
\bottomrule
\end{tabular}
}
%\end{sc}
\end{small}
\end{center}
%\vskip -0.1in
\end{table}

%We also develop different variants of our 4-bit optimized kernels based on the availability of Arm's advanced SIMD vectorization (Neon), scalable vector (SVE) extensions, and the presence or absence of support for matrix-matrix multiply (MMLA) operations in a CPU architecture.
%We also develop different variants of our 4-bit optimized kernels based on their use of various vectorization techniques, such as Arm's advanced SIMD vectorization (Neon) or scalable vector (SVE) extensions, as well as advanced matrix-matrix multiply (MMLA) operations when designing a CPU kernel. Table~\ref{tab:q4_0_4_8_vs_q4_0_4_4_perfcomparison} compares the performance of different variants of our optimized 4-bit group-wise quantized kernels. The Q4\_0\_8\_8 and Q4\_0\_4\_8 kernels are designed with SVE and NEON vector operations, respectively, as well as MMLA operations, whereas the Q4\_0\_4\_4 kernels are designed with only NEON operations and do not include any MMLA instructions. 
We also develop different variants of our $4$-bit optimized kernels based on their weight interleaving patterns, the use of various vectorization techniques, and advanced matrix-matrix multiply (MMLA) operations when designing a CPU kernel. Table~\ref{tab:q4_0_4_8_vs_q4_0_4_4_perfcomparison} compares the performance of different variants of our optimized $4$-bit group-wise quantized kernels. The Q4\_0\_8\_8 and Q4\_0\_4\_8 kernels are designed with MMLA operations, whereas the Q4\_0\_4\_4 kernels do not include any MMLA instructions.
The performance of LLaMA models with Q4\_0\_4\_8 kernels is measured on a Graviton4 processor with $64$ Neoverse V2 cores, while Q4\_0\_4\_4 kernels without MMLA operations are run on a Graviton2 processor with $64$ Neoverse N1 cores that lack MMLA support.

The design of Q4\_0\_4\_8 and Q4\_0\_8\_8 kernels has good similarity; the difference primarily lies in the %use of either NEON or SVE vectorization technology.
weight interleaving patterns between them.
%number of consecutive channels being interleaved.
In the case of the Q4\_0\_4\_8 and Q4\_0\_8\_8 kernels, we perform SIMD-aware weight packing from consecutive four and eight channels, respectively, and the subsequent vector operations on them produce the results of four and eight output channels at once. 
%The weight interleaving patterns between them differ slightly. 
To create an interleaved weight layout, Q4\_0\_8\_8 interleaves eight channels in a group of eight bytes or $16$ $4$-bit quantized elements from each, whereas Q4\_0\_4\_8 interleaves four channels in a group of eight bytes. The improved throughput for Q4\_0\_4\_8 in Table~\ref{tab:q4_0_4_8_vs_q4_0_4_4_perfcomparison} in comparison to Q4\_0\_8\_8 in Table~\ref{tab:q4_0_vs_q4_0_8_8_perfcomparison_grav3} is primarily attributed to changes in Neoverse V2 cores relative to Neoverse V1 and increased memory bandwidth of Graviton4 processors. Q4\_0\_4\_4 interleaves four channels in a group of four bytes. It is worth noting that even without the use of advanced matrix multiply MMLA operations, Q4\_0\_4\_4 achieves significantly better performance in comparison to LLaMA.cpp’s reference Q4\_0 kernels.

%\begin{table*}[!t]
\begin{table}[t]
   \caption{Comparison of the prefill rate and token generation throughput of the LLaMA-3 8B and LLaMA-3.2 3B models for the reference 4-bit Q4\_0 quantization kernel and our optimized Q4\_0\_4\_4, Q4\_0\_8\_8, and Q4\_0\_4\_8 kernels on Arm Graviton2, Graviton3, and Graviton4 CPUs (64 cores), respectively.}
\label{tab:q4_0_4_8_vs_q4_0_4_4_perfcomparison}
%\vskip 0.15in
\begin{center}
\begin{small}
%\begin{sc}
%\begin{tabular}{|c|c|c|c|c|c|}
\scalebox{0.9}{
\begin{tabular}{crrcrrcrrcrr}
\toprule
% & \multicolumn{11}{c}{LLaMA-3 8B}\\
% \cline{2-12}
& \multicolumn{5}{c}{LLaMA-3 8B} & & \multicolumn{5}{c}{LLaMA-3.2 3B}\\
\cline{2-6}
\cline{8-12}
\\
Batch & \multicolumn{2}{c}{Q4\_4\_8 (Optimized)} & & \multicolumn{2}{c}{Q4\_0\_4\_4 (Optimized)} & & \multicolumn{2}{c}{Q4\_0\_4\_8 (Optimized)} & & \multicolumn{2}{c}{Q4\_0\_4\_4 (Optimized)} \\
Size & Prefill & Token Gen. & & Prefill & Token Gen. & & Prefill & Token Gen. & & Prefill & Token Gen.\\
\cline{2-3}
\cline{5-6}
\cline{8-9}
\cline{11-12}
%\cmidrule{2-3}
\\
1 & 643.6 & 66.2 & & 339.6 & 29.1 & & 1210.0 & 115.5 & & 700.8 & 57.1\\
4 & 756.8 & 192.9 & & 355.4 & 92.6 & & 1609.8 & 323.8 & & 784.3 & 161.7\\
8 & 779.5 & 246.2 & & 359.2 & 123.6 & & 1751.4 & 436.7 & & 816.3 & 231.1\\
16 & 767.9 & 341.4 & & 357.8 & 181.1 & & 1739.5 & 568.7 & & 811.5 & 311.2\\
32 & 767.3 & 375.7 & & 353.6 & 186.0 & & 1721.8 & 642.0 & & 800.1 & 325.4\\
\bottomrule
\end{tabular}
}
%\end{sc}
\end{small}
\end{center}
%\vskip -0.1in
\end{table}

\begin{figure}[t]
\centering
    %\includegraphics[width=0.8\textwidth]{figures/token_gen_throughput_comparison-crop.pdf}%\vspace{-3mm}
    % \subfigure[]{\includegraphics[width=0.24\textwidth]{figures/prefill_throughput_comparison_redmi-crop.pdf}}
    % \subfigure[]{\includegraphics[width=0.24\textwidth]{figures/prefill_throughput_comparison_redmi-crop.pdf}}
    \begin{subfigure}[t]{0.5\textwidth}
        \centering
        \includegraphics[width=\linewidth]{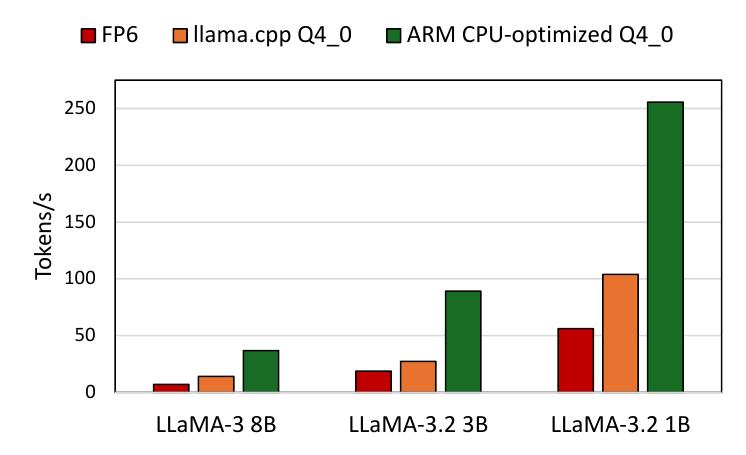}
        \caption{Prefill stage throughput on Redmi K60.}
    \end{subfigure}%
    %\quad
    \begin{subfigure}[t]{0.5\textwidth}
        \centering
        \includegraphics[width=\linewidth]{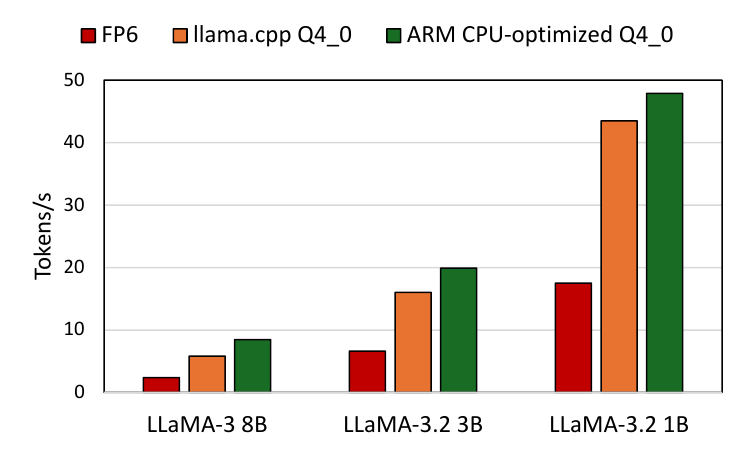}
        \caption{Token generation throughput on Redmi K60.}
    \end{subfigure}
\caption{Comparison of the prefill rate and token generation throughput of the LLaMA-3 8B, LLaMA-3.2 3B, and LLaMA-3.2 1B models on the Redmi K60 mobile device powered by Arm Cortex-series CPUs (4 cores). The batch size is set to one for inference on the mobile device.}
\label{fig:redmik60_throughput_comparison}
\end{figure}

Figure~\ref{fig:redmik60_throughput_comparison} compares the prefill rate and token generation throughput of LLaMA models with parameter sizes ranging from 1B to 8B on the Redmi K60 mobile device, which is powered by four Arm Cortex-series CPU cores. For inference on mobile devices, the batch size is set to one. For the LLaMA-3.2 1B parameter model, we improve the inference speed for the prefill phase from $104$ tokens/s to $255.6$ tokens/s and the token generation phase from $43.4$ tokens/s to $47.8$ tokens/s through our $4$-bit optimized kernels.

%\begin{table*}[!t]
\begin{table}[t]
   \caption{Comparison of the prefill rate and token generation throughput of the LLaMA-3 8B and LLaMA-3.2 3B models for the reference llama.cpp 4-bit non-uniform quantization kernel (IQ4\_NL) and our corresponding optimized kernel on Arm Graviton3 CPUs (64 cores).}
\label{tab:iq4_nl_vs_iq4_nl_n_perfcomparison}
%\vskip 0.15in
\begin{center}
\begin{small}
%\begin{sc}
%\begin{tabular}{|c|c|c|c|c|c|}
\scalebox{0.9}{
\begin{tabular}{crrcrrcrrcrr}
\toprule
& \multicolumn{5}{c}{LLaMA-3 8B} & & \multicolumn{5}{c}{LLaMA-3.2 3B}\\
\cline{2-6}
\cline{8-12}
\\
Batch & \multicolumn{2}{c}{IQ4\_NL (llama.cpp)} & & \multicolumn{2}{c}{IQ4\_NL (Optimized)} & & \multicolumn{2}{c}{IQ4\_NL (llama.cpp)} & & \multicolumn{2}{c}{IQ4\_NL (Optimized)} \\
Size & Prefill & Token Gen. & & Prefill & Token Gen. & & Prefill & Token Gen. & & Prefill & Token Gen.\\
\cline{2-3}
\cline{5-6}
\cline{8-9}
\cline{11-12}
%\cmidrule{2-3}
\\
1 & 132.6 & 45.5 & & 470.1 & 45.5 & & 282.2 & 83.2 & & 1015.7 & 87.6 \\
4 & 138.3 & 90.1 & & 520.5 & 142.0 & & 327.8 & 180.3 & & 1274.0 & 264.7 \\
8 & 145.7 & 105.0 & & 531.1 & 184.8 & & 356.1 & 217.7 & & 1357.3 & 356.4\\
16 & 145.5 & 114.0 & & 527.7 & 282.9 & & 355.6 & 242.0 & & 1349.5 & 506.3\\
32 & 145.0 & 118.5 & & 522.8 & 304.9 & & 351.7 & 254.6 & & 1333.6 & 554.0\\
\bottomrule
\end{tabular}
}
%\end{sc}
\end{small}
\end{center}
%\vskip -0.1in
\end{table}

\subsection{Inference throughput for 4-bit non-uniform quantization}
We extend our proposed SIMD-aware weight packing and fast decompression path optimizations to $4$-bit non-uniform codebook-based quantization methods, such as LLaMA.cpp's IQ4\_NL~\cite{llamacpp}. IQ4\_NL employs a single $16$-entry $8$-bit integer codebook with a non-uniform distribution, similar to a Normal Float-~\cite{dettmers2023qlora} or Student Float-like~\cite{dotzel2024sf4} normal distribution, to map $4$-bit quantized indices into $8$-bit integer values. In addition to the aforementioned optimizations, our optimized IQ4\_NL kernel for Arm CPUs can take advantage of existing vector table lookup instructions (\texttt{vtbl}). 
\texttt{vtbl} can perform a vector read to access multiple byte values at once corresponding to quantized indexes in the input vector from a codebook table during inference. Because IQ4\_NL only has a single $16$-entry codebook, all of its entries can fit in a vector register and be accessed from there during inference. This, in turn, enables faster throughput in both the compute-bound prefill and small-batch-sized memory-bound token generation stages when compared to the LLaMA.cpp's reference IQ4\_NL kernel, as illustrated in Table~\ref{tab:iq4_nl_vs_iq4_nl_n_perfcomparison}. Furthermore, the inference speed of our optimized IQ4\_NL on Graviton3 CPUs in Table~\ref{tab:iq4_nl_vs_iq4_nl_n_perfcomparison} is comparable to that of the optimized Q4\_0\_8\_8 kernel in Table~\ref{tab:q4_0_vs_q4_0_8_8_perfcomparison_grav3}.
This also confirms the seamless support of off-the-shelf Arm CPUs in efficiently running non-uniform codebook-based quantization methods along with uniform quantization.

\subsection{Inference throughput for narrower 2-bit quantization}

% Figure~\ref{fig:token_gen_throughput_varying_threads} compares the runtime performance of Q2\_NL to the uniform and non-uniform quantization 
% techniques, Q2\_K and IQ2\_S, respectively. We report the performance of the autoregressive decode 
% stage in LLM inference for a single thread. For a single thread count, LLM inference is compute bound.
% To better understand the comparative performance of the various 2-bit schemes, we normalized the 
% numbers to the performance (tokens/sec) of the 4-bit quantization scheme, Q4\_0. For compute-bound 
% prefill (time-to-first-token) and autoregressive decoding stages with fewer thread counts, the 
% decompression overhead for a 2-bit quantization scheme in converting 2-bit quantized weights to 8-bit 
% values should be low enough to have a comparable throughput to that of an 8-bit quantized scheme or a 
% 4-bit scheme with comparatively low dequantization overhead (for example, Q4\_0). Because of its high 
% dequantization overhead, the non-uniform IQ2\_S performs poorly in comparison to the Q4\_0's 
% throughput, as shown in Table 2. In contrast, the uniform Q2\_K quantization method has a lower 
% dequantization overhead and helps bridge the performance gap with Q4\_0.

\begin{figure}[tb]
\centering
\includegraphics[width=0.8\textwidth]{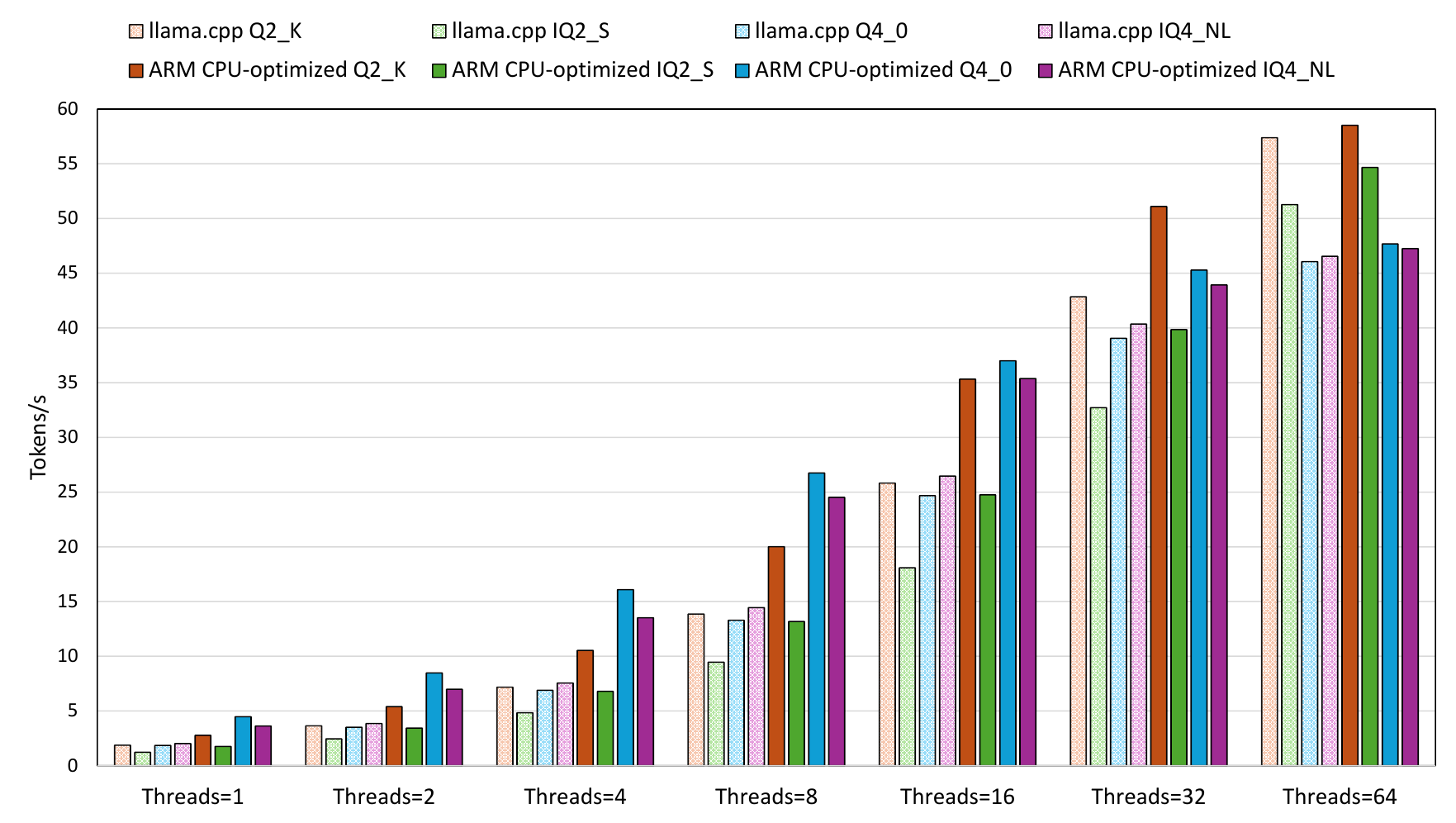}%\vspace{-3mm}
	\caption{Comparison of the token generation throughput of the LLaMA-3 8B model for different quantization schemes of varying bit-widths. Performance was measured for different numbers of threads (cores) on Arm Graviton3 CPUs.}
\label{fig:token_gen_throughput_varying_threads}
\vspace{-4mm}
\end{figure}

Figure~\ref{fig:token_gen_throughput_varying_threads} 
shows the runtime performance of our optimized, narrower bit-width group-quantized kernels on Arm CPUs, specifically $2$-bit uniform and non-uniform quantization methods, such as Q2\_K and IQ2\_S from LLaMA.cpp~\cite{llamacpp}. Q2\_K is a group-wise $2$-bit uniform quantization technique, whereas IQ2\_S is a $2$-bit non-uniform codebook-based technique. The IQ2 family of quantization methods from LLaMA.cpp adopts some of the key compression techniques proposed in the QuIP\#~\cite{tseng2024quip} work, especially the E8 lattice-based codebook for vector quantization and its symmetric properties of an even number of positive (or negative) signs in quantized vectors, in conjunction with a group-wise structure over LLM weight matrices. We report the performance of the token generation stage in LLM inference for varying thread counts. For a small number of threads, the token generation stage is compute bound. To better understand the comparative performance of the various $2$-bit schemes, we compare them to the runtime performance (tokens/sec) of $4$-bit quantization schemes. For the compute-bound token generation stage with lower thread counts, as well as the prefill stage, the decompression overhead for a $2$-bit quantization scheme in converting $2$-bit quantized weights to $8$-bit values should be low enough to avoid dequantization becoming a bottleneck.
%For compute-bound prefill and token generation stages with fewer thread counts, the decompression overhead for a 2-bit quantization scheme in converting 2-bit quantized weights to 8-bit values should be low enough to have a comparable throughput to that of a 4-bit quantized scheme with comparatively low dequantization overhead (for example, Q4\_0). 
%The uniform Q2\_K quantization method has a lower dequantization overhead and helps bridge the performance gap of Arm CPU-optimized Q2\_K with Arm CPU-optimized Q4\_0\_8\_8 for lower thread counts. For the memory-bound token generation phase at large thread counts, the decompression overhead does not play a role in actual speedup, and the low memory bandwidth pressure of 2-bit quantization methods assists in achieving better throughout than 4-bit methods.
The uniform Q2\_K quantization method has a low dequantization overhead, which helps bridge the performance gap between Arm CPU-optimized Q2\_K and Arm CPU-optimized Q4\_0\_8\_8 at lower thread counts. For the memory-bound token generation phase at higher thread counts, the decompression overhead has no effect on actual speedup, and the low memory bandwidth pressure of $2$-bit quantization methods helps to achieve better overall performance than $4$-bit methods. On the other hand, the non-uniform codebook-based method, IQ2\_S, incurs a high dequantization overhead in constructing codebook indices from compressed weights and then accessing codebooks, so even our highly optimized IQ2\_S (Arm CPU-optimized IQ2\_S) performs poorly when compared to the optimized Q2\_K's (Arm CPU-optimized Q2\_K) throughput at lower thread counts, as shown in Figure~\ref{fig:token_gen_throughput_varying_threads}.

The poor performance of existing codebook-based methods for low bit-widths like IQ2\_S motivates the development of our group-wise fine-grained codebook-based quantization, a simple yet effective method for low-bit non-uniform quantization. Our group-wise codebook-based quantization is built around fast inference, not only to ensure high throughput during memory-bound token generation phases at higher thread counts but also to minimize dequantization overhead and achieve high throughput during compute-bound LLM inference phases at lower thread counts while achieving better accuracy.
%Our group-wise codebook-based quantization is optimized for fast inference, ensuring high throughput not only during memory bound token generation phase at higher thread counts, but also with low dequantization overhead and high throughput during compute bound LLM inference phases at lower thread counts.

\subsection{Accuracy evaluation for group-wise codebook-based quantization}

%We examine the accuracy (perplexity) of our generated models on the WikiText2 dataset, which is standard in this setting.
We evaluate the quantized models on token perplexity for the WikiText2 validation set. Table~\ref{tab:wikitext2_ppl_comparison} reports the perplexity metric numbers for WikiText2. 
We compare our group-wise non-uniform codebook-based quantization approach
%Q2\_NL 
against several recent state-of-the-art post-training uniform and non-uniform quantization methods targeting LLMs, including the IQ2 and IQ3 family of models, Q2K from LLaMA.cpp~\cite{llamacpp}, and %QuIP\#~\cite{tseng2024quip} 
SqueezeLLM~\cite{kim2024sqllm}
on the LLaMA family of models with varying parameters~\cite{touvron2023llama2openfoundation}. As previously mentioned, the IQ2 and IQ3 quantization  approaches closely adopt the quantization techniques from the state-of-the-art QuIP\#~\cite{tseng2024quip}.
Additional accuracy results for other models, centroid value distributions of different codebooks,  and ablation studies for group-wise codebook quantization are provided in the Appendix.
%The IQ2 family of quantization methods from LLaMA.cpp adopts some of the key compression techniques proposed in the QuIP\#~\cite{tseng2024quip} work, especially the E8 lattice-based codebook for vector quantization and its symmetric properties of an even number of positive (or negative) signs in quantized vectors, in conjunction with a group-wise structure over LLM  weight matrices.

Activations are quantized using $8$-bit group-wise uniform quantization across all experiments for various weight quantization methods. For each LLM layer, the Q2\_K quantization method quantizes the majority of weight matrices to $2$-bit while a few of them to $3$-bit.
%\begin{table*}[!t]
\begin{table}[t]
   \caption{Comparison of the perplexity score on WikiText2 for a sequence length of $512$. Results for SqueezeLLM~\cite{kim2024sqllm} were obtained using the released codebase for LLaMA.cpp. 2-bit quantization is unsupported by the SqueezeLLM codebase.}
\label{tab:wikitext2_ppl_comparison}
%\vskip 0.15in
\begin{center}
\begin{footnotesize}
%\begin{sc}
%\begin{tabular}{|c|c|c|c|c|c|}
\scalebox{0.9}{
%\begin{tabular}{lcccccccccr}
\begin{tabular}{lcccc}
\toprule
Method & Avg. bits per weight & Quantization & LLaMA-2 7B & LLaMA-3 8B\\% & LLaMA-3.2 3B\\
& (activations: $8$-bit) & type & &\\
\midrule
FP16 & 16 &  & 5.79 & 6.23\\
\midrule
%RTN & - & Uniform &  &  &  &\\
%GPTQ & - & Uniform &  &  &  &\\
%SmoothQuant & - & Uniform &  &  &  &\\
SqueezeLLM & 4.04 & Non-uniform & 5.97 & - \\
\midrule
Q8\_0 & 8.50 & Uniform & 5.80 & 6.23\\
Q6\_K & 6.56 & Uniform & 5.81 & 6.25\\
Q5\_K\_M & 5.70 & Uniform & 5.83 & 6.29\\
Q5\_0 & 5.58 & Uniform & 5.83 & 6.36\\
Q4\_K\_M & 4.89 & Uniform & 5.87 & 6.38\\
Q4\_0 & 4.65 & Uniform & 5.96 & 6.70\\
IQ4\_NL & 4.65 & Non-uniform & 5.87 & 6.45\\
Q3\_K\_M & 3.99 & Uniform & 6.00 & 6.73\\
IQ3\_M & 3.76 & Non-uniform & 6.02 & 6.89\\
IQ3\_XS & 3.49 & Non-uniform & 6.12 & 7.16\\
Q3\_K\_S & 3.64 & Uniform & 6.21 & 7.60\\
Q2\_K & 3.15 & Uniform & 6.68 & 8.65\\
Q2\_K\_S & 2.96 & Uniform & 7.21 & 9.32\\
IQ2\_M & 2.92 & Non-uniform & 6.59 & 8.60\\
IQ2\_S & 2.74 & Non-uniform & 7.01 & 9.65\\
IQ2\_XS & 2.58 & Non-uniform & 7.52 & 10.76\\
\textbf{Group-wise codebook} & 3.2 & Non-uniform & \textbf{6.39} & \textbf{7.77}\\
%codebook (ours) & & & & & &\\
\bottomrule
\end{tabular}
}
%\end{sc}
\end{footnotesize}
\end{center}
\vskip -0.1in
\end{table}
To ensure a fair comparison with a uniform quantization technique such as Q2\_K with similar bits per weight, all $2$-bit and $3$-bit uniform quantized weight matrices of it are compressed using group-wise codebook-based quantization. For both $2$-bit and $3$-bit quantized layers, a handful of $4$-entry and $8$-entry $8$-bit codebooks, respectively, are found using Algorithm~\ref{alg:q2nl_quantization} and used during PTQ. 
Furthermore, there is no need to determine the codebook entries for a new, unseen LLM.  For accuracy evaluation in Table~\ref{tab:wikitext2_ppl_comparison}, the codebooks found for the LLaMA2 7B model weights using Algorithm~\ref{alg:q2nl_quantization} are used during PTQ for the LLaMA3 8B model. This ensures the generalization performance of codebooks found using Algorithm~\ref{alg:q2nl_quantization} after scaling weight values to codebook bit-width via group-wise quantization. It is worth noting that, while the majority of the codebooks found for $2$-bit and $3$-bit quantized layers using Algorithm~\ref{alg:q2nl_quantization} capture mostly symmetric distributions of various shapes, a few also capture asymmetric distributions found in group-wise quantized LLM weights.
As shown in Table~\ref{tab:wikitext2_ppl_comparison}, our fine-grained codebook-based quantization technique outperforms both the most effective uniform quantization technique, Q2\_K, and the non-uniform quantization technique, IQ2\_S, in terms of perplexity while requiring similar bits per weight. For the LLaMA3 8B model, the uniform quantization method Q2\_K achieves a perplexity of $8.65$, whereas our group-wise codebook quantization offers a better perplexity of $7.77$ at similar bits per weight ($3.15-3.2$ bits).

The fewer number of $4$- or $8$-entry codebooks for our group-wise codebook quantization ensures that all the codebooks can fit in the vector register file of Arm CPU cores during the course of inference as opposed to fitting in L1 cache in case of IQ2\_S-like techniques. 
In other words, there is no need to load the codebook entries to the register file for different weight rows or LLM layers; once loaded from memory, the register file can store the entire codebook. The same codebook essentially applies to all layers of an LLM.
In contrast, for IQ2\_S-like techniques using E8 lattice-based codebooks from QuIP\#~\cite{tseng2024quip}, while the entire codebook can fit in L1 cache, it cannot fit in the register file of CPU cores to enable fast access to codebooks.
For our group-wise codebook quantization, the codebook index for a group and the centroid indices that specify particular centroid values from the assigned codebook can be combined using bitwise vector operations before using specialized vector table lookup operations \texttt{vtbl} to retrieve centroid values from register file resident codebooks.
The simple dequantization path, akin to Q2\_K-like uniform quantization techniques, and low overhead in accessing register file resident codebooks using \texttt{vtbl} in our group-wise codebook-based quantization ensures fast inference and comparable throughput to that of Q2\_K, whereas the fine-grained assignment of codebooks to each group in a weight tensor ensures better accuracy than Q2\_K.
For the memory-bound decode stage with large thread counts (for example, $64$ threads), both quantization schemes, Q2\_K and our group-wise codebook-based quantization, can achieve higher throughput due to their reduced model footprint and the consequent ease of memory bandwidth, as observed in our experiments. It is worth noting that, like IQ2, the 3-bit IQ3 quantization schemes have a high decompression overhead when creating codebook indices and accessing codebooks, resulting in poor runtime performance.

To summarize, our group-wise, non-uniform codebook-based quantization scheme not only outperforms the state-of-the-art in terms of quality by matching the underlying distribution of weight values with codebooks, but it also ensures comparable throughput performance to an equivalent uniform quantization scheme with low decompression overhead for both compute-bound and memory-bound stages in LLM inference. In other words, it presents a pareto-optimal solution in terms of model quality and runtime performance for narrow bit-widths such as $2$-bit and $3$-bit quantization.
While we apply our fine-grained codebook-based quantization technique to $2$ and $3$ bit-widths, it can also be used to find group-wise codebooks for other bit-widths, such as $4$-bit quantization.

%% file: sec/conclusion.tex
\section{Conclusion}
\label{sec:conclusion}
In this work, we propose highly optimized matrix multiply kernels to accelerate LLM inference on CPUs. Our optimized kernels for Arm CPU-based inference for LLaMA models can deliver a $3$-$3.2\times$ improvement in prompt processing and an approximately $2\times$ improvement in token generation runtime for $4$-bit quantized LLMs, significantly enhancing both memory efficiency and inference speed over the best existing LLaMA.cpp-based open-source solution for CPUs. Furthermore, we present highly efficient kernels for $4$-bit non-uniformly quantized and narrower $2$-bit uniformly and non-uniformly quantized LLMs, demonstrating the efficacy of off-the-shelf CPUs, particularly Arm CPUs, in supporting non-uniform and narrower bit-width quantizations. However, the complex decompression path of existing non-uniform codebook-based quantization schemes for narrower bit-widths, such as $2$ bits per weight, presents a significant challenge to achieving good runtime performance even with sophisticated kernel optimizations. We address this through the introduction of a group-wise fine-grained codebook-based quantization scheme, which can better match the different non-uniform distribution patterns existing across different groups in LLM weights. It outperforms state-of-the-art non-uniform and uniform quantization methods, achieving higher accuracy at similar bits per weight while potentially ensuring significantly faster runtime than non-uniform quantization methods and comparable runtime to light-weight uniform quantization methods. There are other group-wise quantization schemes for various bit-widths, and the kernel optimizations and fine-grained codebooks proposed here should be extended to other schemes as well.